\documentclass[twoside]{article}
\usepackage{aistats2015}
\usepackage{times,url,mathrsfs}
\usepackage{amsmath,wrapfig,color}
\usepackage{amsfonts}
\usepackage{graphicx}
\usepackage{subfigure}
\usepackage{bbm}
\newtheorem{theorem}{Theorem}

\begin{document}

%

%

\twocolumn[

\aistatstitle{\vspace{-0.15in}Improved Asymmetric Locality Sensitive Hashing (ALSH) for Maximum Inner Product Search (MIPS)\vspace{-0.15in}}

\vspace{-0.2in}

\aistatsauthor{Anshumali Shrivastava \And Ping Li  }

\aistatsaddress{ Department of Computer Science \\ Computing and Information Science\\ Cornell University, Ithaca, NY 14853, USA\\\texttt{anshu@cs.cornell.edu}
 \And Department of Statistics and Biostatistics \\ Department of Computer Science\\  Rutgers University, Piscataway, NJ 08854, USA \\\texttt{pingli@stat.rutgers.edu}} ]

\begin{abstract}
\vspace{-0.2in}
Recently it was shown that the problem of Maximum Inner Product Search (MIPS) is efficient and it admits provably sub-linear hashing algorithms. Asymmetric transformations before hashing were the key in solving MIPS which was otherwise hard. In~\cite{Report:ALSH_arXiv14}, the authors use asymmetric transformations which convert the problem of approximate MIPS into the problem of approximate near neighbor search which can be efficiently solved using hashing. In this work, we provide a different transformation which converts the problem of approximate MIPS into the problem of approximate cosine similarity search which can be efficiently solved using signed random projections.  Theoretical analysis show that the new scheme is significantly better than the original scheme for MIPS. Experimental evaluations strongly support the theoretical findings.
\end{abstract}

\vspace{-0.2in}
\section{Introduction}
\vspace{-0.1in}

In this paper, we revisit the problem of \emph{Maximum Inner Product Search (MIPS)}, which was studied in a recent technical report~\cite{Report:ALSH_arXiv14}. In this report the authors present the first provably fast algorithm for MIPS, which was considered hard~\cite{Proc:Ram_KDD12,Proc:Koenigstein_CIKM12}. Given an input query point
$q \in \mathbb{R}^D$, the task of MIPS is to find $p \in \mathcal{S}$, where $\mathcal{S}$ is a giant collection of size $N$, which maximizes (approximately) the {\bf inner product $q^Tp$}:
\begin{equation}
p = \arg\max_{x \in \mathcal{S}}\hspace{0.1in} q^Tx
\end{equation}
The MIPS problem is related to the problem of \emph{ near neighbor search (NNS)}. For example, L2-NNS
\begin{equation}
p = \arg\min_{x \in \mathcal{S}}||q -x||_2^2 =  \arg\min_{x \in \mathcal{S}} (||x||_2^2 -  2 q^Tx)
\end{equation}
or, correlation-NNS
\begin{equation}
p = \arg\max_{x \in \mathcal{S}}\frac{q^Tx}{\|q\|\|x\|} =  \arg\max_{x \in \mathcal{S}} \frac{q^Tx}{\|x\|}
\end{equation}

These three problems are equivalent if the norm of every element $x \in \mathcal{S}$ is constant. Clearly, the value of the norm $||q||_2$ has no effect for the argmax.  In many scenarios, MIPS arises naturally at places where the norms of the elements in $\mathcal{S}$ have  significant variations~\cite{Proc:Koenigstein_CIKM12}. As reviewed in~\cite{Report:ALSH_arXiv14}, examples of applications of MIPS include recommender system~\cite{Article:Koren_2009,Proc:Cremonesi_RecSys,Proc:Koenigstein_CIKM12},   large-scale object detection with DPM~\cite{Article:Felzenszwalb_PAMI2010,Proc:Dean_CVPR2013,Article:joachims_2009,Article:joachims_2009}, structural SVM~\cite{Proc:Dean_CVPR2013}, and  multi-class label prediction~\cite{Proc:Ram_KDD12,Proc:Koenigstein_CIKM12,Proc:Weber_VLDB98}.

\noindent\textbf{Asymmetric LSH (ALSH)}: Locality Sensitive Hashing (LSH)~\cite{Proc:Indyk_STOC98} is popular in  practice for efficiently solving NNS.  In the prior work~\cite{Report:ALSH_arXiv14}, the concept of ``asymmetric LSH'' (ALSH) was proposed that one can transform the input query $Q(p)$ and data in the collection $P(x)$ independently, where the transformations $Q$ and $P$ are different. \cite{Report:ALSH_arXiv14} developed a particular set of transformations to convert MIPS into L2-NNS and then solved the problem by standard L2-hash~\cite{Proc:Datar_SCG04}. In this paper,  we name the scheme in~\cite{Report:ALSH_arXiv14} as \textbf{L2-ALSH}. Asymmetry in hashing has become popular recently, and it has been applied for hashing higher order similarity~\cite{Proc:Shrivastava_NIPS13}, data dependent hashing~\cite{Proc:Neyshabur_NIPS13}, sketching~\cite{Proc:Dong_SIGIR08} etc.

\noindent\textbf{Our contribution}: In this study, we propose another scheme for ALSH, by developing a new set of asymmetric transformations to convert  MIPS into a problem of correlation-NNS, which is solved by ``sign random projections''~\cite{Article:Goemans_JACM95,Proc:Charikar}. We name this new scheme as \textbf{Sign-ALSH}. Our theoretical analysis and experimental study show that Sign-LSH is more advantageous than L2-ALSH for MIPS.

\vspace{-0.15in}
\section{Review: Locality Sensitive Hashing (LSH)}
\vspace{-0.1in}

The problem of efficiently finding nearest neighbors has been an active research since the very early days of computer science~\cite{Article:Friedman_74}.
Approximate versions of the near neighbor search problem~\cite{Proc:Indyk_STOC98} were proposed to break the linear query time bottleneck. The following formulation for approximate near neighbor search is often adopted.

\noindent\textbf{Definition:} ($c$-Approximate Near Neighbor or $c$-NN)\ {\em Given a set of points in a $D$-dimensional space $\mathbb{R}^D$, and parameters $S_0 > 0$, $\delta > 0$, construct a data structure which, given any query point $q$, does the following with probability $1- \delta$: if there exists an $S_0$-near neighbor of $q$ in $\mathcal{S}$, it reports some $cS_0$-near neighbor of $q$ in $\mathcal{S}$.}

\emph{Locality Sensitive Hashing} (LSH)~\cite{Proc:Indyk_STOC98} is a family of functions, with the property that more similar items have a higher  collision probability. LSH trades off query time with extra (one time) preprocessing  cost and space.  Existence of an LSH family translates into provably sublinear query time algorithm for c-NN problems.

\noindent\textbf{Definition:} (Locality Sensitive Hashing (LSH))\ {\em A family $\mathcal{H}$ is called $(S_0,cS_0,p_1,p_2)$-sensitive if, for any two points $x,y \in \mathbb{R}^D$, $h$ chosen uniformly from $\mathcal{H}$ satisfies:
\begin{itemize}
\vspace{-0.1in}
\item if $Sim(x,y)\ge S_0$ then $Pr_\mathcal{H}(h(x) = h(y)) \ge p_1$
\item if $ Sim(x,y)\le cS_0$ then $Pr_\mathcal{H}(h(x) = h(y)) \le p_2$
\vspace{-0.1in}
\end{itemize}
For efficient approximate nearest neighbor search, $p_1 > p_2$ and $c < 1$ is needed.}

\noindent\textbf{Fact 1}:\ Given a family of $(S_0,cS_0,p_1,p_2)$ -sensitive hash functions, one can construct a data structure for $c$-NN with $O(n^\rho \log{n})$ query time and space $O(n^{1 + \rho})$, where $\rho = \frac{\log{p_1}}{\log{p_2}} <1$.

LSH is a generic framework and an implementation of LSH requires  a concrete hash function.

\vspace{-0.05in}
\subsection{LSH for L2 distance}\label{sec:L2Hash}
\vspace{-0.05in}
\cite{Proc:Datar_SCG04} presented an LSH family for $L_2$ distances.  Formally, given a fixed window size $r$, we sample  a random vector $a$ with each component  from i.i.d. normal, i.e., $a_i \sim N(0,1)$, and a scalar $b$ generated uniformly at random from $[0,r]$. The hash function  is defined as:
\begin{equation}\label{eq:L2Hash} h_{a,b}^{L2}(x) = \left\lfloor \frac{a^Tx + b}{r} \right\rfloor \end{equation}
where $\lfloor \rfloor$ is the floor operation. The collision probability under this scheme can be shown to be
\begin{align}\label{eq:L2collprob}
&Pr(h^{L2}_{a,b}(x) = h^{L2}_{a,b}(y)) \\ \notag
&=1 - 2\Phi(-r/d) - \frac{2}{\sqrt{2\pi}(r/d)}\left(1 - e^{-(r/d)^2/2}\right)
\end{align}
where $\Phi(x) = \int_{-\infty}^{x}\frac{1}{\sqrt{2\pi}}e^{-\frac{x^2}{2}}dx$  and  $d = ||x -y||_2$ is the Euclidean distance between the vectors $x$ and $y$.

\vspace{-0.05in}
\subsection{LSH for correlation}
\vspace{-0.05in}

Another popular LSH family is the so-called ``sign random projections''~\cite{Article:Goemans_JACM95,Proc:Charikar}.  Again, we choose  a random vector $a$ with  $a_i \sim N(0,1)$. The hash function  is defined as:
\begin{equation}\label{eq:SignHash}
h^{Sign}(x) = sign(a^Tx)
\end{equation}
And collision probability is
\begin{align}\label{eq:Signcollprob}
&Pr(h^{Sign}(x) = h^{Sign}(y)) =1 -\frac{1}{\pi}\cos^{-1}\left(\frac{x^Ty}{\|x\|\|y\|}\right)
\end{align}
This hashing scheme is also popularly known as \emph{signed random projections (SRP)}
\section{Review of ALSH  for MIPS and L2-ALSH}

In~\cite{Report:ALSH_arXiv14}, it was shown that the framework of locality sensitive hashing is restrictive for solving MIPS. The inherent assumption of the same  hash function for both the transformation as well as the query was unnecessary in the classical LSH framework and it was the main hurdle in finding provable sub-linear algorithms for MIPS with LSH.  For the theoretical guarantees of LSH to work there was no requirement of symmetry.  Incorporating asymmetry in the hashing schemes was the key in solving MIPS efficiently.

\noindent\textbf{Definition~\cite{Report:ALSH_arXiv14}:} ({\bf\it Asymmetric} Locality Sensitive Hashing (ALSH))\  A family $\mathcal{H}$, along with the two vector functions  $Q:\mathbb{R}^D \mapsto \mathbb{R}^{D'}$ ({\bf Query Transformation}) and $P:\mathbb{R}^D \mapsto \mathbb{R}^{D'}$ ({\bf Preprocessing Transformation}), is called $(S_0,cS_0,p_1,p_2)$-sensitive if for a given $c$-NN instance with query $q$, and the hash function $h$ chosen uniformly from $\mathcal{H}$ satisfies the following:
\begin{itemize}
\vspace{-0.1in}
\item if $Sim(q,x)\ge S_0$ then $Pr_\mathcal{H}(h(Q(q))) = h(P(x))) \ge p_1$
\item if $ Sim(q,x)\le cS_0$ then $Pr_\mathcal{H}(h(Q(q)) = h(P(x))) \le p_2$
\vspace{-0.1in}
\end{itemize}
Here $x$ is any point in the collection $\mathcal{S}$.

Note that the query transformation $Q$ is only applied on the query and the pre-processing transformation $P$ is applied to $x \in \mathcal{S}$ while creating hash tables.   By letting $Q(x) = P(x) = x$,  we can recover the vanilla LSH. Using different transformations (i.e., $Q \ne P$), it is possible to counter the fact that self similarity is not highest with inner products which is the main argument of failure of LSH. We only just need the probability of the new collision event $\{ h(Q(q)) = h(P(y))\}$ to satisfy the conditions of definition of ALSH for $Sim(q,y) = q^Ty$.

\begin{theorem}~\cite{Report:ALSH_arXiv14}\label{theo:extendedLSH}
 Given a family of hash function $\mathcal{H}$ and the associated query and preprocessing transformations $P$ and $Q$, which is $(S_0,cS_0,p_1,p_2)$ -sensitive, one can construct a data structure for $c$-NN with $O(n^\rho \log{n})$ query time and space $O(n^{1 + \rho})$, where $\rho = \frac{\log{p_1}}{\log{p_2}}$.
\vspace{-0.05in}
\end{theorem}

\cite{Report:ALSH_arXiv14} also provided an explicit construction of ALSH, which we call \textbf{ L2-ALSH}. Without loss of generality, one can always assume
\begin{align}
||x_i||_2 \le U < 1,\ \ \forall x_i \in \mathcal{S}
\end{align}
for some $U < 1$. If this is not the case, then we can always scale down the norms without altering the $\arg\max$.
Since the norm of the query does not affect the $\arg\max$ in MIPS, for simplicity it was assumed $||q||_2 = 1$. This condition can be removed easily (see Section~\ref{sec:Remove} for details).
In L2-ALSH,  two vector transformations $P:\mathbb{R}^D \mapsto \mathbb{R}^{D+m}$ and $Q:\mathbb{R}^D \mapsto  \mathbb{R}^{D+m}$ are defined as follows:
\begin{align}
\label{eq:P}P(x) &= [x; ||x||^2_2;||x||^4_2; ....;||x||^{2^m}_2]\\
\label{eq:Q}Q(x) &= [x; 1/2; 1/2; ....; 1/2],
\end{align}
where [;] is the concatenation. $P(x)$ appends $m$ scalers of the form $||x||_2^{2^i}$ at the end of the vector $x$, while Q(x) simply appends $m$ ``1/2'' to the end of the vector $x$.  By observing
\begin{align}\notag
&||P({x_i})||^2_2 = ||x_i||^2_2 + ||x_i||^4_2+ ... + ||x_i||^{2^m}_2 +  ||x_i||^{2^{m+1}}_2\\\notag
&||Q({q})||^2_2= ||q||^2_2 + m/4 = 1+ m/4\\
&Q(q)^TP({x_i}) = q^Tx_i + \frac{1}{2}( ||x_i||^2_2 + ||x_i||^4_2+ ... + ||x_i||^{2^m}_2) \notag
\end{align}
 one can obtain the following key equality:
\begin{equation}\label{eq:errorterminEquation}||Q(q) - P({x_i})||^2_2 = (1 +  m/4) -  2q^Tx_i  +||x_i||^{2^{m+1}}_2\end{equation}
Since $||x_i||_2 \le U < 1$, we have $||x_i||^{2^{m+1}}  \rightarrow 0$
at the tower rate (exponential to exponential). Thus,  as long as $m$ is not too small (e.g., $m\geq3$ would suffice), we have
\begin{equation}
\label{eq:MIPSNNS}
\arg\max_{x \in \mathcal{S}} q^Tx  \simeq \arg\min_{x \in \mathcal{S}} ||Q(q) - P({x})||_2
\end{equation}
This scheme is the first connection between solving un-normalized MIPS and approximate near neighbor search. Transformations $P$ and $Q$, when norms are less than 1, provide correction to the L2 distance $||Q(q) - P({x_i})||_2$ making it rank correlate with the (un-normalized) inner product.   The general idea of ALSH was partially inspired by the work on three-way similarity search~\cite{Proc:Shrivastava_NIPS13}, where they applied different hashing functions for handling query and data in the repository.

\begin{figure}[t]
\vspace{-0.2in}
\begin{center}
\mbox{
\includegraphics[width=2.7in]{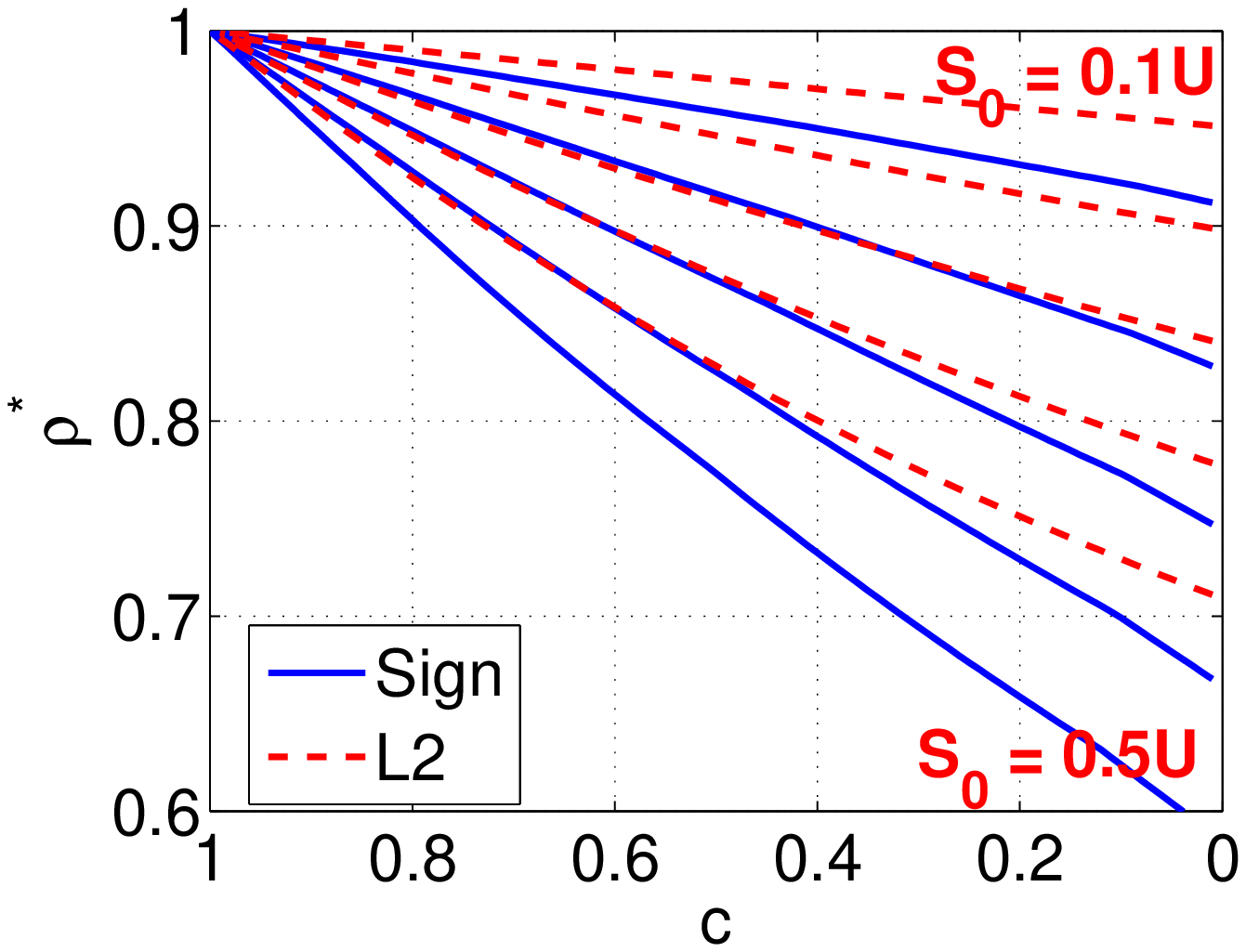}
}

\mbox{
\includegraphics[width=2.7in]{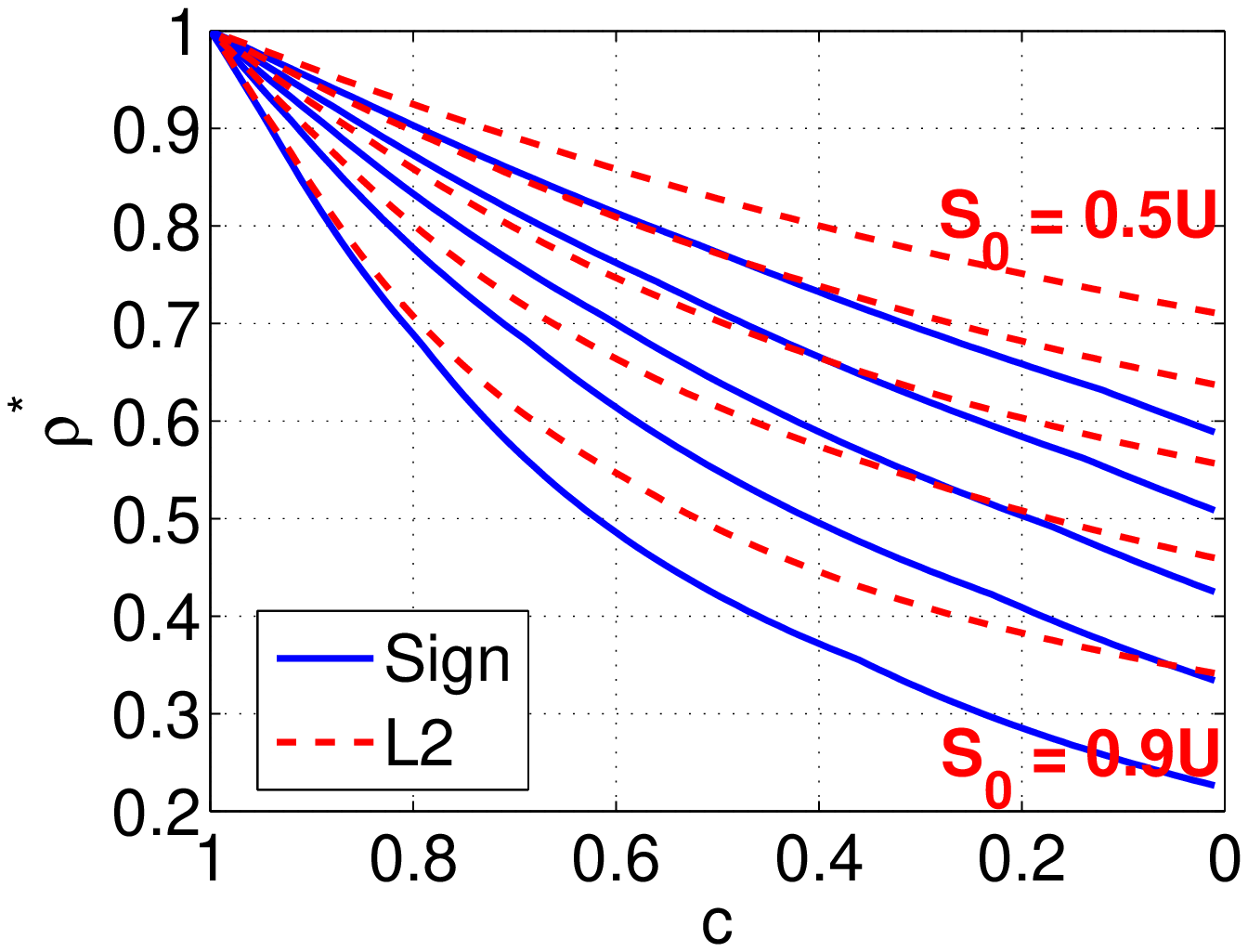}
}
\end{center}
\vspace{-0.2in}
\caption{Optimal values of $\rho^*$ (lower is better) with respect to approximation ratio $c$ for different $S_0$, obtained by a grid search over parameters $U$ and $m$, given $S_0$ and $c$. The curves show that Sign-ALSH (solid curves) is noticeably better than L2-ALSH (dashed curves) in terms of their optimal $\rho^*$ values. The results for L2-ALSH were from the prior work~\cite{Report:ALSH_arXiv14}. For clarity, the results are in two figures.   }\label{fig:OptRho}\vspace{-0.1in}
\end{figure}

 \begin{figure}[t]
\vspace{-0.2in}
\begin{center}
\mbox{
\includegraphics[width=2.7in]{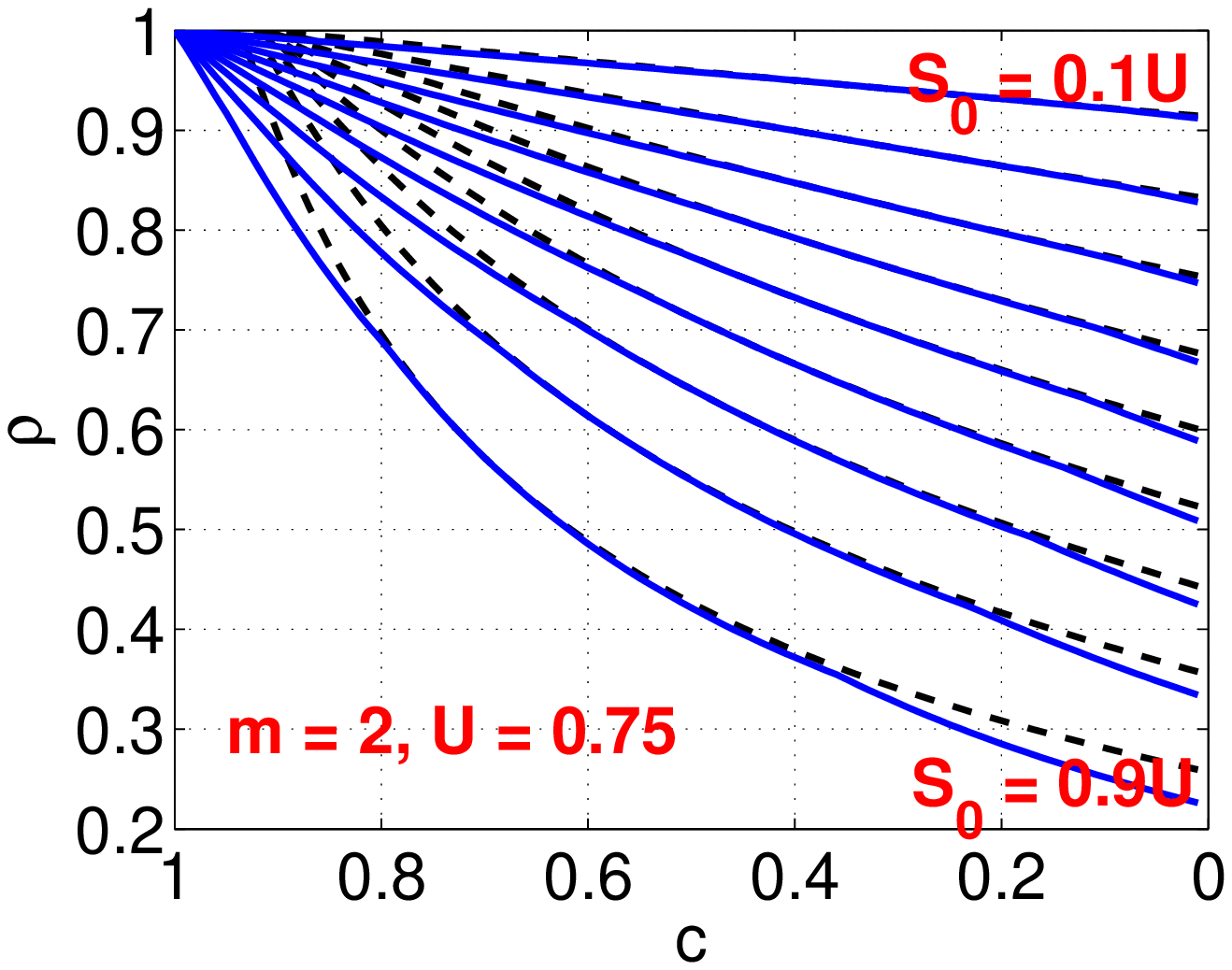}
}
\mbox{
\includegraphics[width=2.7in]{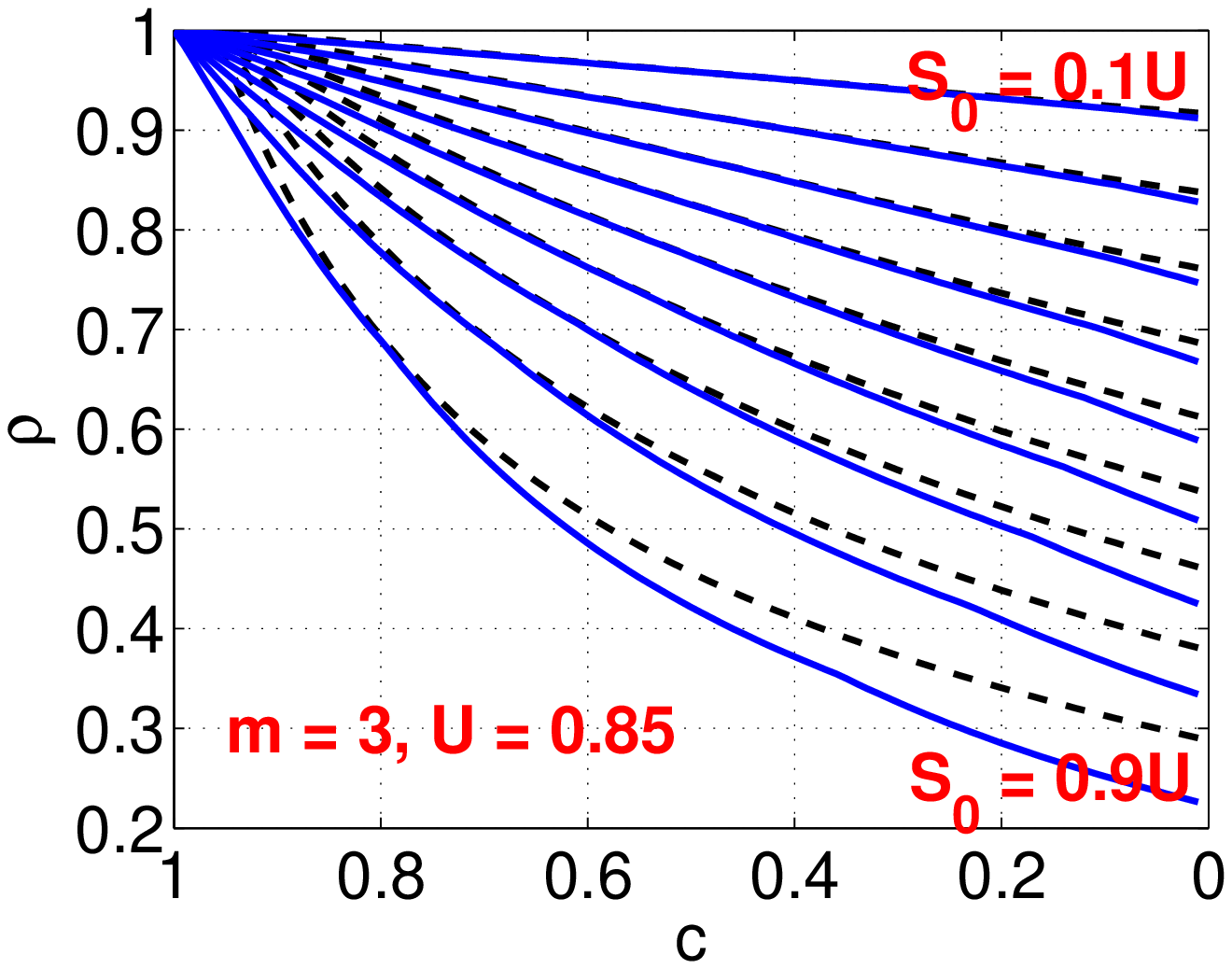}
}
\end{center}
\vspace{-0.2in}
\caption{The solid curves are the optimal $\rho$ values of Sign-ALSH from Figure~\ref{fig:OptRho}. The dashed curves represent the $\rho$ values for fixed parameters: $m=2$ and $U=0.75$ (left panel), and $m=3$ and $U=0.85$ (right panel). Even with fixed parameters, the $\rho$ does not degrade much.}\label{fig:FixedRho}\vspace{-0.2in}
\end{figure}

\vspace{-0.05in}
\subsection{Intuition for the Better Scheme }
\vspace{-0.05in}
Asymmetric transformations give us enough flexibility to modify norms without changing inner products. The transformation provided in~\cite{Report:ALSH_arXiv14} used this flexibility to convert MIPS to  standard near neighbor search in $L_2$ space for which we have standard hash functions. Signed random projections are popular hash functions widely adopted for correlation or cosine similarity. We use asymmetric transformation to convert approximate MIPS into approximate maximum correlation search.  The transformations and the collision probability of the hashing functions determines the efficiency of the  obtained ALSH algorithm. We show that the new transformation with SRP is better suited for ALSH compared to the existing L2-ALSH. Note that in the recent work on {\em coding for random projections}~\cite{Proc:Li_ICML14,Report:RPCodeLSH2014}, it was already shown that sign random projections (or 2-bit random projections) can outperform L2LSH.

\vspace{-0.1in}
\section{The New Proposal: Sign-ALSH}
\vspace{-0.1in}

\subsection{From MIPS to Correlation-NNS}
\label{sec:construction}
\vspace{-0.05in}

We assume for simplicity that $||q||_2 = 1$ as the norm of the query does not change the ordering, we show in the next section how to get rid of this assumption. Without loss of generality let $||x_i||_2 \le U < 1,\ \ \forall x_i \in \mathcal{S}$ as it can always be achieved by scaling the data by large enough number.  We define two vector transformations $P:\mathbb{R}^D \mapsto \mathbb{R}^{D+m}$ and $Q:\mathbb{R}^D \mapsto  \mathbb{R}^{D+m}$ as follows:
\begin{align}
\label{eq:P}P(x) &= [x; 1/2-||x||^2_2;1/2-||x||^4_2; ....;1/2-||x||^{2^m}_2]\\
\label{eq:Q}Q(x) &= [x; 0; 0; ....; 0],
\end{align}
Using $||Q({q})||^2_2= ||q||^2_2=1$,  $Q(q)^TP({x_i}) = q^Tx_i$, and
\begin{align}\notag
\vspace{-0.1in}
&||P({x_i})||^2_2 \\\notag
&= ||x_i||^2_2 + 1/4+||x_i||^4_2 -||x_i||^2_2 + 1/4+||x_i||^8_2 -||x_i||^4_2+ ... \\\notag
&+ 1/4+||x_i||^{2^{m+1}}_2 - ||x_i||^{2^{m}}_2\\\notag
&=m/4 + ||x_i||^{2^{m+1}}_2
\end{align}
 we obtain the following key equality:
\begin{align}\label{eq:errorterminEquation}
\frac{Q(q)^TP({x_i})}{{\|Q(q)\|_2\|P({x_i})\|_2}} = \frac{q^Tx_i}{\sqrt{m/4+||x_i||^{2^{m+1}}_2 }}
\end{align}
The term $||x_i||^{2^{m+1}}  \rightarrow 0,$ again vanishes  at the tower rate. This means we have approximately
\begin{equation}
\label{eq:MIPSNNS}
\arg\max_{x \in \mathcal{S}} q^Tx  \simeq \arg\max_{x \in \mathcal{S}} \frac{Q(q)^TP({x_i})}{{\|Q(q)\|_2\|P({x_i})\|_2}}
\end{equation}
This  provides another solution for solving  MIPS using known methods for  approximate correlation-NNS.

\vspace{-0.05in}
\subsection{Fast Algorithms for MIPS Using Sign Random Projections}
\vspace{-0.05in}
Eq. (\ref{eq:MIPSNNS}) shows that MIPS reduces to the standard approximate near neighbor search problem which can be efficiently solved by sign random projections, i.e., $h^{Sign}$ (defined by Eq. (\ref{eq:SignHash})).  Formally, we can state the following theorem.
\begin{theorem}
\vspace{-0.1in}
\label{theo:collprobnew}
Given a $c$-approximate  instance of MIPS, i.e., $Sim(q,x) = q^Tx$, and a query $q$ such that $||q||_2 = 1$ along with a collection $\mathcal{S}$ having $||x||_2 \le  U <1$  $\forall x \in \mathcal{S}.$ Let $P$ and $Q$ be the vector transformations defined in Eq. (\ref{eq:P}) and Eq. (\ref{eq:Q}), respectively.  We have the following two conditions for hash function $h^{Sign}$ (defined by Eq. (\ref{eq:SignHash}))
\begin{itemize}
\vspace{-0.1in}
\item if $q^Tx \ge S_0$ then
\begin{align}
\notag &Pr[h^{Sign}(Q(q)) = h^{Sign}(P(x))] \\\notag &\ge 1-\frac{1}{\pi}\cos^{-1}\left(\frac{S_0}{\sqrt{m/4 + U^{2^{m+1}}}}\right)
\end{align}
\vspace{-0.3in}
\item if $ q^Tx \le cS_0$ then
\begin{align}\notag
&Pr[h^{Sign}(Q(q)) = h^{Sign}(P(x))] \\\notag &\le 1-\frac{1}{\pi}\cos^{-1}\left(\frac{\min\{cS_0,z^*\}}{\sqrt{m/4 + \left(\min\{cS_0,z^*\}\right)^{2^{m+1}}}}\right)
\end{align}
where $z^*=\left(\frac{m/2}{2^{m+1}-2}\right)^{2^{-m-1}}$.
\vspace{-0.1in}
\end{itemize}
\noindent\textbf{Proof:}\ When $q^Tx \geq S_0$,  we have, according to Eq. (\ref{eq:Signcollprob})
\begin{align}\notag
 &Pr[h^{Sign}(Q(q)) = h^{Sign}(P(x))] \\
\notag&=1-\frac{1}{\pi}\cos^{-1}\left(\frac{q^Tx}{\sqrt{m/4+||x||_2^{2^{m+1}}}}\right) \\ \notag&\geq 1-\frac{1}{\pi}\cos^{-1}\left(\frac{q^Tx}{\sqrt{m/4+U^{2^{m+1}}}}\right)
\end{align}
When $q^Tx\leq cS_0$, by noting that $q^Tx \leq \|x\|_2$,  we have
\begin{align}\notag
 &Pr[h^{Sign}(Q(q)) = h^{Sign}(P(x))]\\
\notag &=1-\frac{1}{\pi}\cos^{-1}\left(\frac{q^Tx}{\sqrt{m/4+||x||_2^{2^{m+1}}}}\right) \\\notag & \leq 1-\frac{1}{\pi}\cos^{-1}\left(\frac{q^Tx}{\sqrt{m/4+(q^Tx)^{2^{m+1}}}}\right)
\end{align}
For this one-dimensional function $f(z) = \frac{z}{\sqrt{a+z^b}}$, where $z=q^Tx$, $a=m/4$ and $b = 2^{m+1}\geq2$, we know
\begin{align}\notag
f^\prime(z) = \frac{a-z^b\left(b/2-1\right)}{(a+z^b)^{3/2}}
\end{align}
One can also check that $f^{\prime\prime}(z) \leq 0$ for $0<z<1$, i.e., $f(z)$ is a concave function. The maximum of $f(z)$ is attained at $z^* =\left( \frac{2a}{b-2}\right)^{1/b}=\left(\frac{m/2}{2^{m+1}-2}\right)^{2^{-m-1}}$
If $z^* \geq cS_0$, then we need to use $f(cS_0)$ as the bound. $\hfill\Box$
\end{theorem}

Therefore, we have  obtained, in LSH terminology,
\begin{align}
&p_1 = 1-\frac{1}{\pi}\cos^{-1}\left(\frac{S_0}{\sqrt{m/4 + U^{2^{m+1}}}}\right)\\
&p_2 = 1-\frac{1}{\pi}\cos^{-1}\left(\frac{\min\{cS_0,z^*\}}{\sqrt{m/4 + \left(\min\{cS_0,z^*\}\right)^{2^{m+1}}}}\right),\hspace{0.2in}\\& z^*=\left(\frac{m/2}{2^{m+1}-2}\right)^{2^{-m-1}}
\end{align}
Theorem~\ref{theo:extendedLSH} allows us to construct data structures with  worst case $O(n^\rho \log{n})$ query time guarantees for $c$-approximate MIPS, where $\rho = \frac{\log p_1}{\log p_2}$.  For any given $c <1$, there always exist $U <1$ and $m$ such that $\rho < 1$.  This way, we obtain a sublinear query time algorithm for MIPS.  Because $\rho$ is a function of 2 parameters, the best query time chooses $U$ and $m$, which minimizes the value of $\rho$. For convenience, we define
\begin{align}\label{eq:optrho}
\rho^* = \min_{U,m} \frac{\log\left(1-\frac{1}{\pi}\cos^{-1}\left(\frac{S_0}{\sqrt{m/4 + U^{2^{m+1}}}}\right)\right)}{\log\left(1-\frac{1}{\pi}\cos^{-1}\left(\frac{\min\{cS_0,z^*\}}{\sqrt{m/4 + \left(\min\{cS_0,z^*\}\right)^{2^{m+1}}}}\right)\right)}
\end{align}
See Figure~\ref{fig:OptRho} for the plots of $\rho^*$, which also compares the optimal $\rho$ values for L2-ALSH in the prior work~\cite{Report:ALSH_arXiv14}. The results show that Sign-ALSH is noticeably better.

\vspace{-0.1in}
\subsection{Parameter Selection}\label{sec_parameter}
\vspace{-0.1in}

Figure~\ref{fig:FixedRho} presents the $\rho$ values for two sets of selected parameters: $(m, U)=(2,0.75)$ and $(m, U)=(3,0.85)$. We can see that even if we use fixed parameters, the performance would not degrade much. This essentially frees practitioners from the burden of choosing parameters.

\vspace{-0.15in}
\section{Remove Dependence on  Norm of  Query}
\label{sec:Remove}
\vspace{-0.1in}

Changing norms of the query does not affect the $\arg\max_{x \in \mathcal{C}} q^Tx$, and hence, in practice for retrieving top-$k$, normalizing the query should not affect the performance. But for theoretical purposes, we want the runtime guarantee to be independent of $||q||_2$.  Note, both LSH and ALSH schemes solve the $c$-approximate instance of the problem, which requires a threshold $S_0 = q^tx$ and an approximation ratio $c$.  For this given $c$-approximate instance we choose optimal parameters $K$ and $L$. If the queries have varying norms, which is likely the case in practical scenarios, then given a $c$-approximate MIPS instance, normalizing the query will change the problem because it will change the threshold $S_0$ and also the approximation ratio $c$. The optimal parameters for the algorithm $K$ and $L$, which are also the size of the data structure, change with $S_0$ and $c$. This will require re-doing the costly preprocessing with every change in query.  Thus, the query time which is dependent on $\rho$ should be independent of the query.

Transformations $P$ and $Q$ were precisely meant to remove the dependency of correlation on the norms of $x$ but at the same time keeping the inner products same. Realizing the fact that we are allowed asymmetry, we can use the same idea to get rid of the norm of $q$. Let $M$ be the upper bound on all the norms i.e. $ M = max_{x \in \mathcal{C}} ||x||_2 $. In other words $M$ is the radius of the space.

Let $U <1$, define the transformations, $T:  \mathbb{R}^D \rightarrow \mathbb{R}^D$ as
\begin{align}
\label{eq:T}T(x) &= \frac{Ux}{M}
\end{align}
and transformations $P, Q: \mathbb{R}^D \rightarrow \mathbb{R}^{D+m}$ are the same for the Sign-ALSH scheme as defined in Eq (\ref{eq:P}) and (\ref{eq:Q}).

Given the query $q$ and any data point $x$, observe that the inner products between $P(Q(T(q)))$ and $Q(P(T(x)))$ is
\begin{align}
P(Q(T(q)))^T Q(P(T(x))) = q^Tx \times\left(\frac{U^2}{M^2}\right)
\end{align}

$P(Q(T(q)))$ appends first m zeros components to $T(q)$ and then $m$ components of the form $1/2 -||q||^{2^i}$. $Q(P(T(q)))$ does the same thing but in a different order. Now we are working in $D +2m$ dimensions. It is not difficult to see that the norms of  $P(Q(T(q)))$ and $Q(P(T(q)))$  is given by
\begin{align}
||P(Q(T(q)))||_2 &= \sqrt{\frac{m}{4} + ||T(q)||_2^{2^{m+1}}}\\
||Q(P(T(x)))||_2 &= \sqrt{\frac{m}{4} + ||T(x)||_2^{2^{m+1}}}
\end{align}
The transformations are very asymmetric but we know that it is necessary.

\begin{figure*}[ht]
\vspace{-0.15in}
\begin{center}
\mbox{
\includegraphics[width=2in]{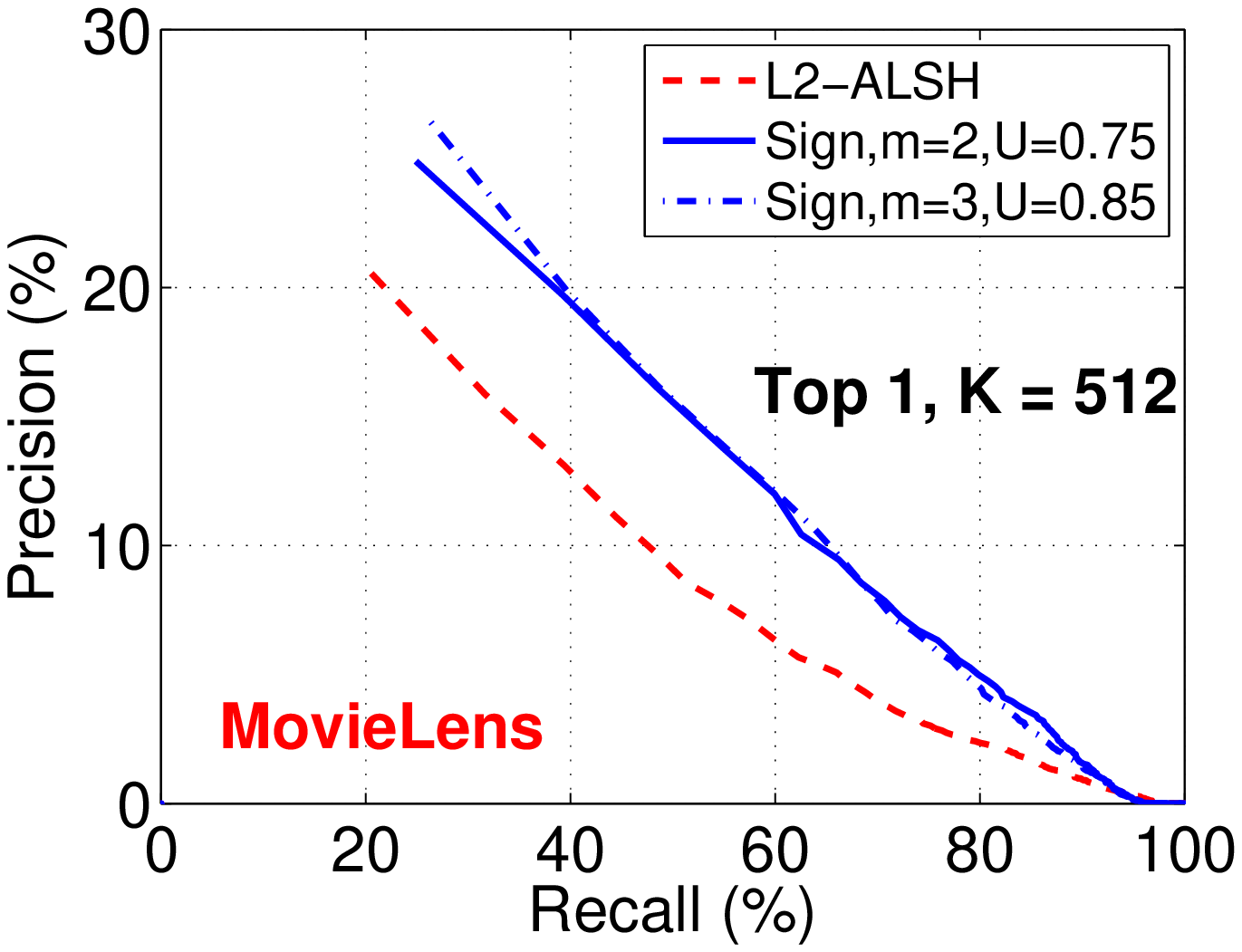}\hspace{0.13in}
\includegraphics[width=2in]{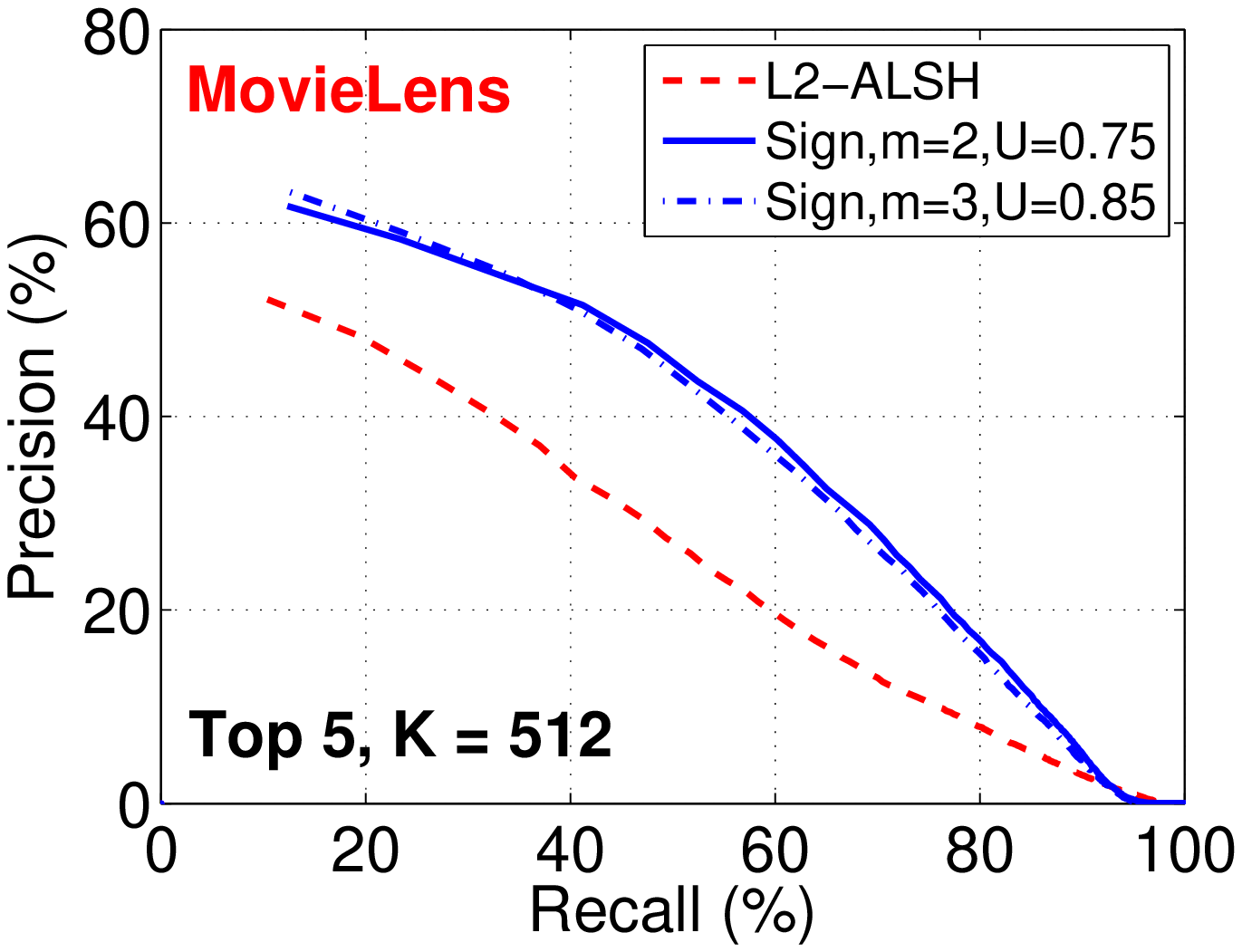}\hspace{0.13in}
\includegraphics[width=2in]{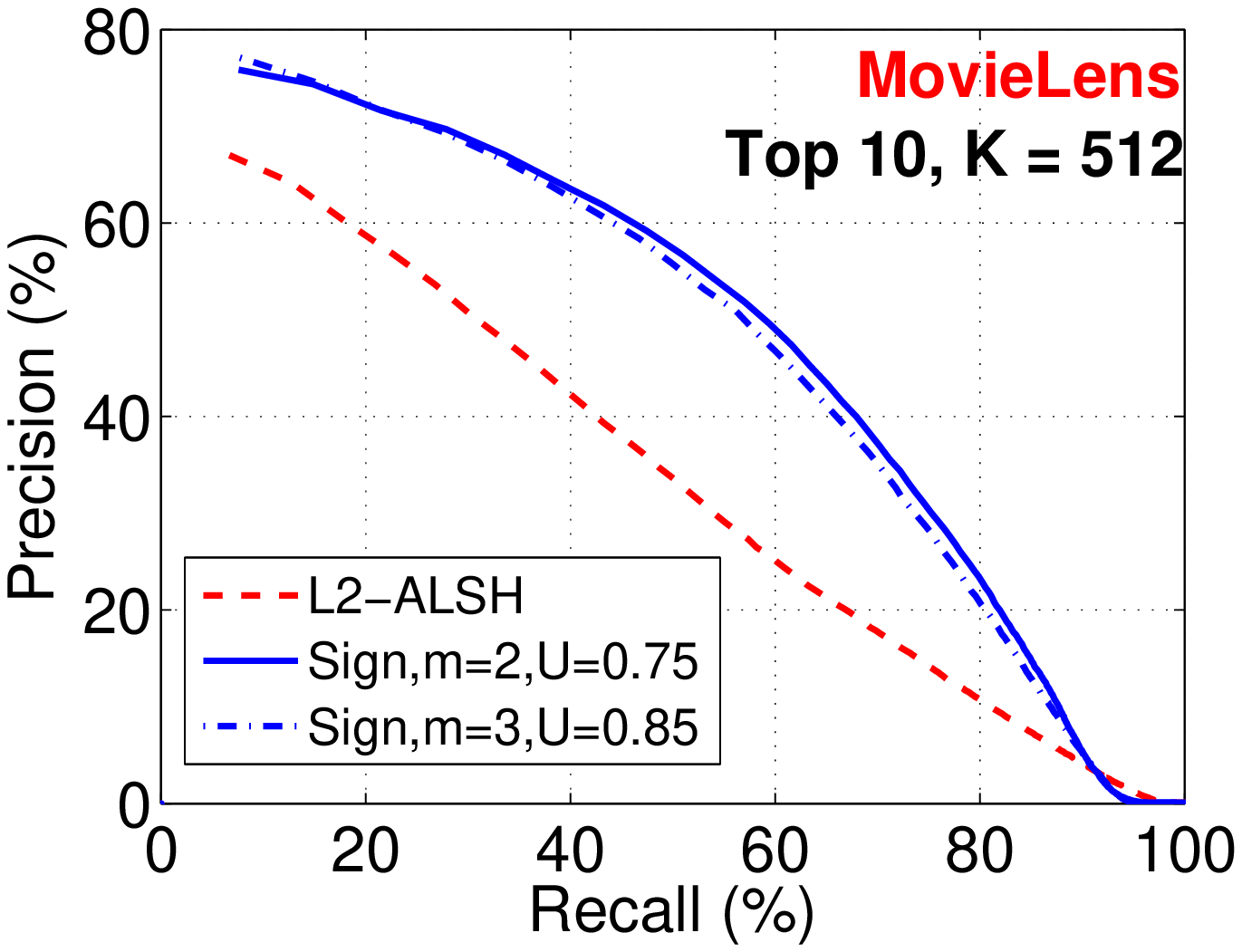}
}

\mbox{
\includegraphics[width=2in]{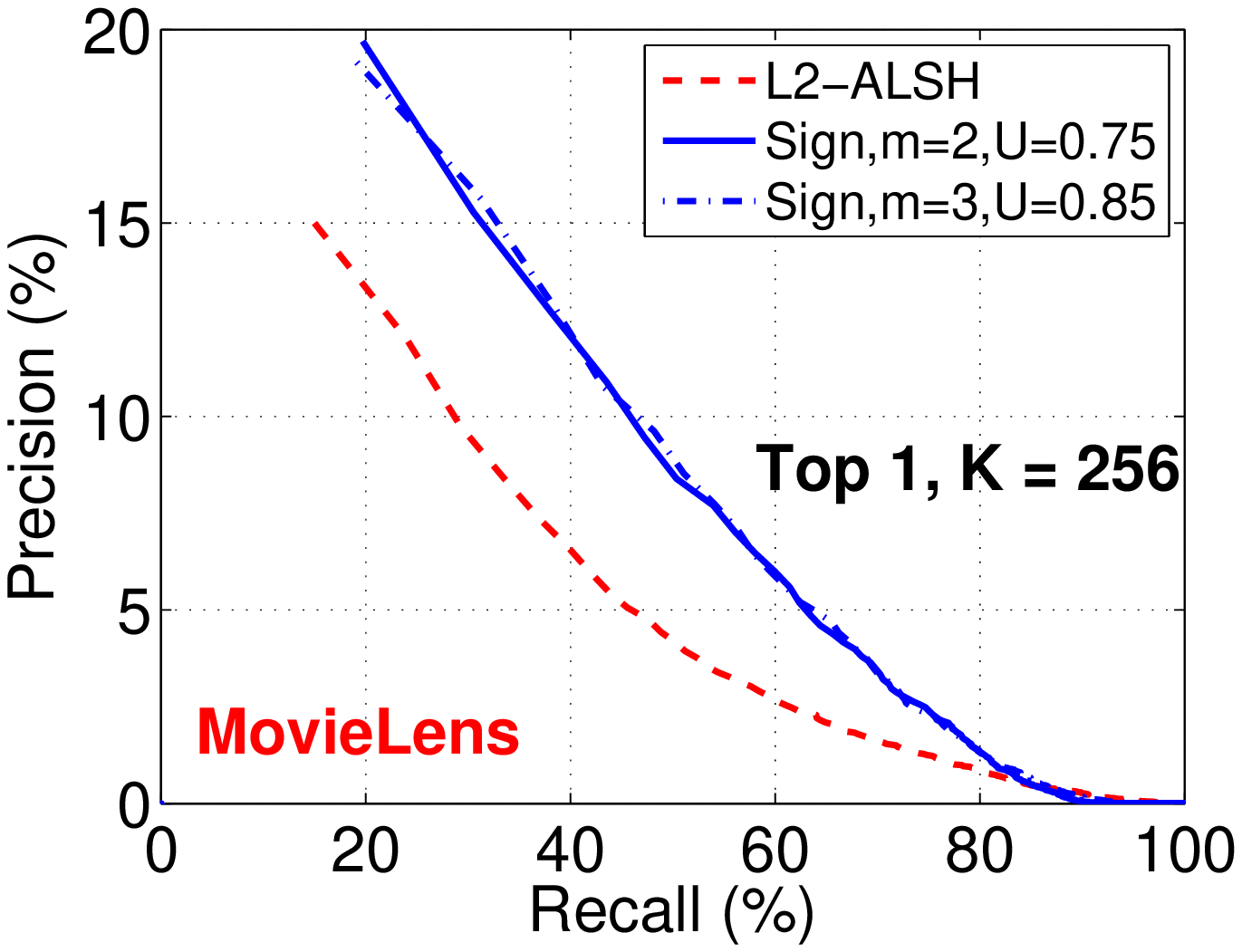}\hspace{0.13in}
\includegraphics[width=2in]{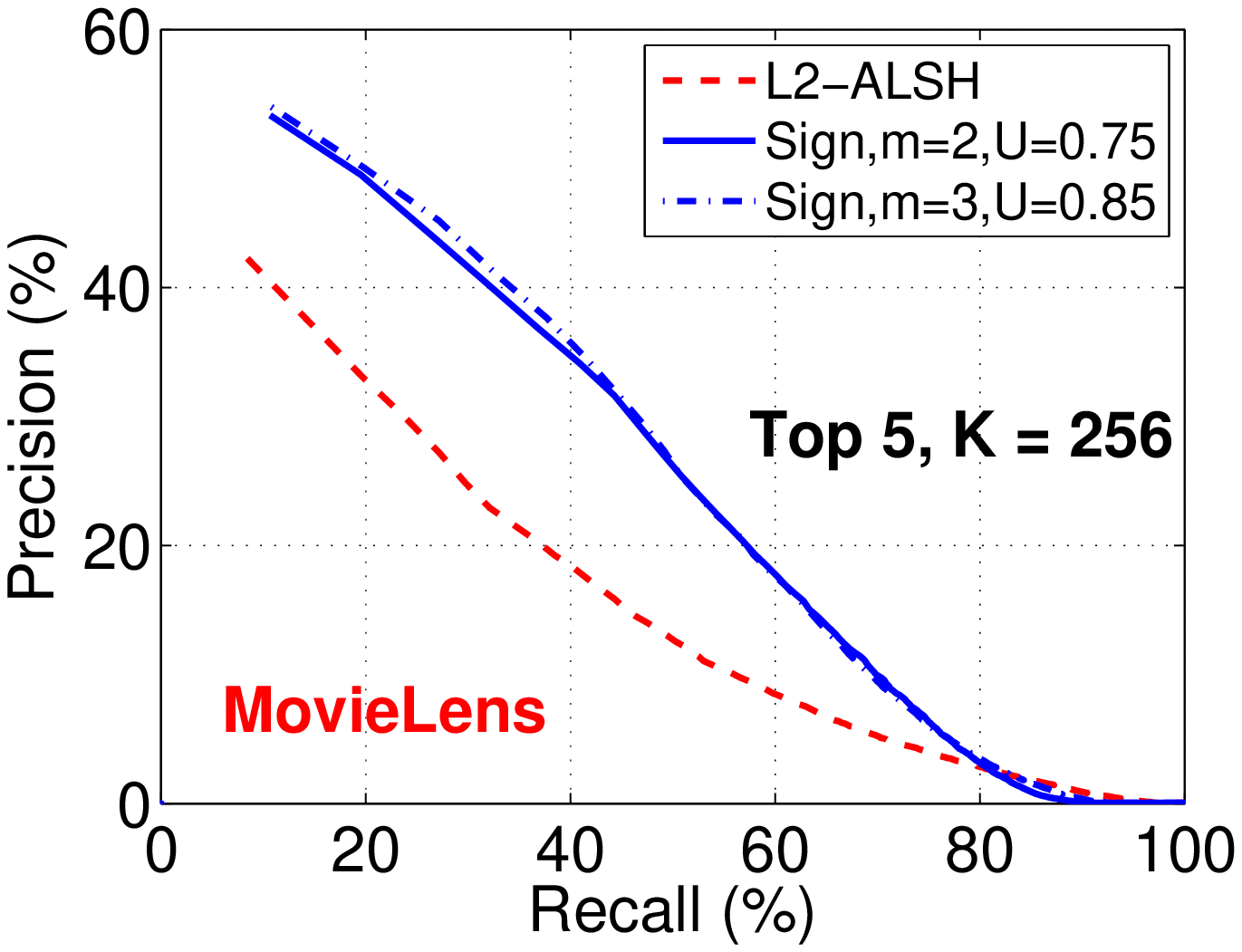}\hspace{0.13in}
\includegraphics[width=2in]{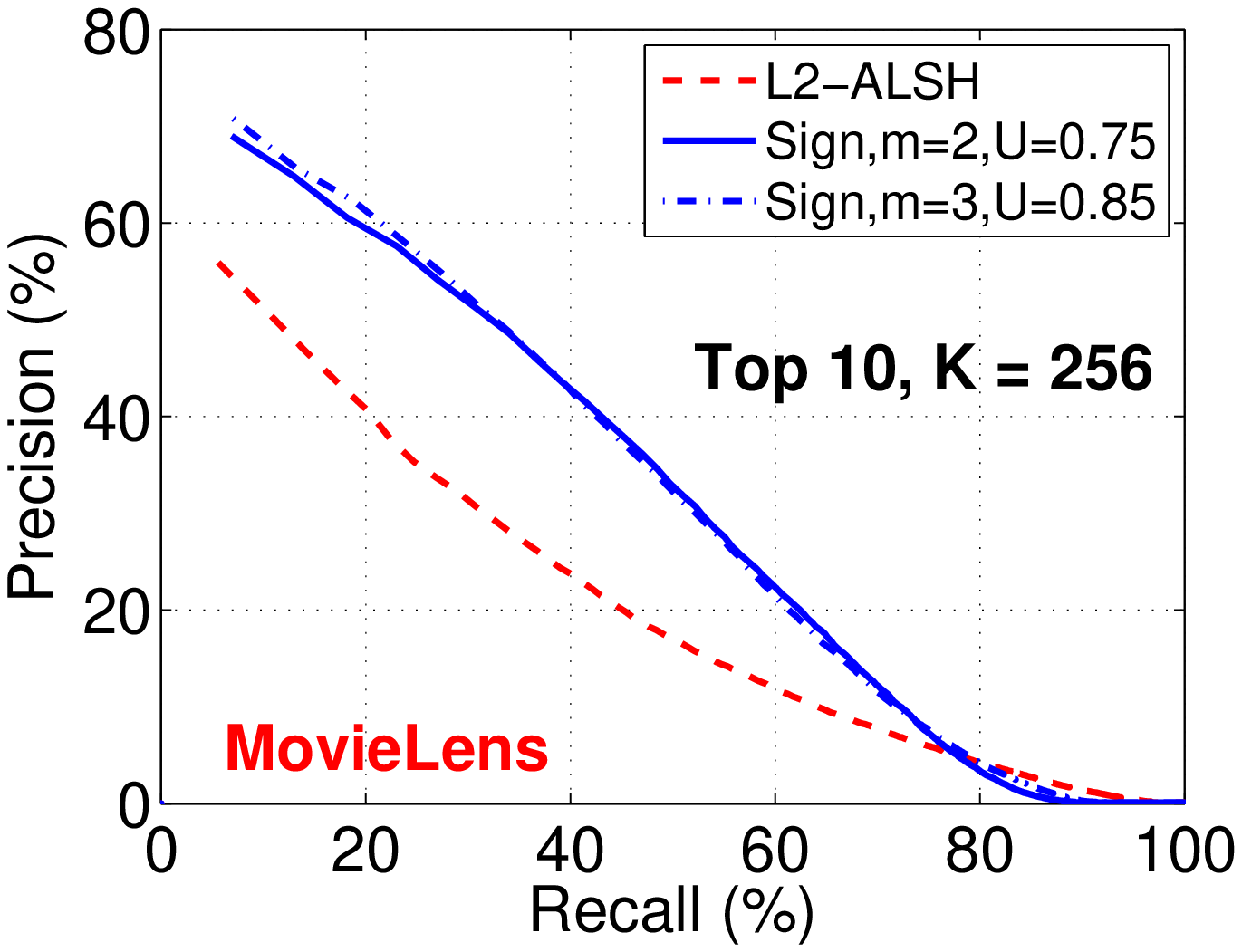}
}

\mbox{
\includegraphics[width=2in]{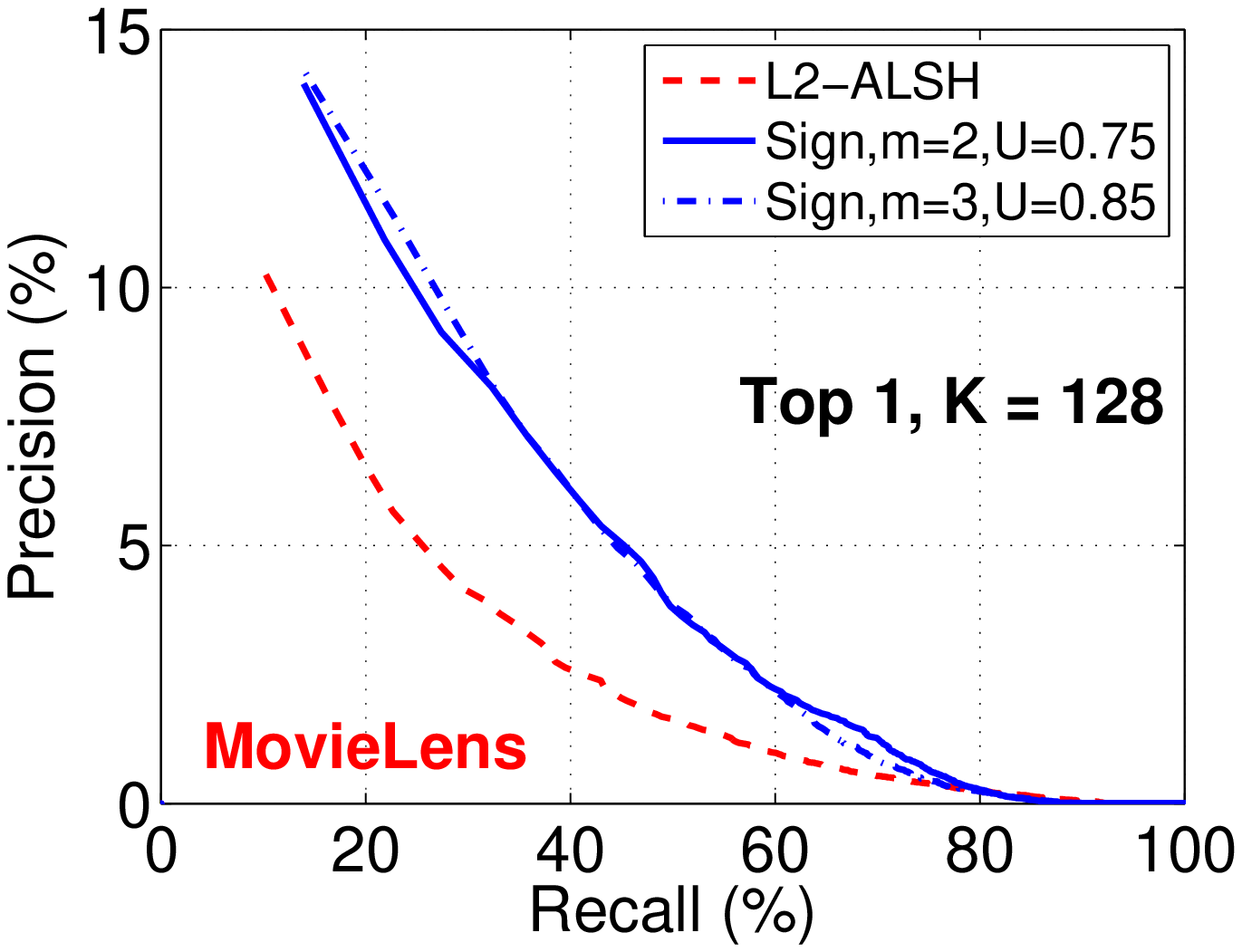}\hspace{0.13in}
\includegraphics[width=2in]{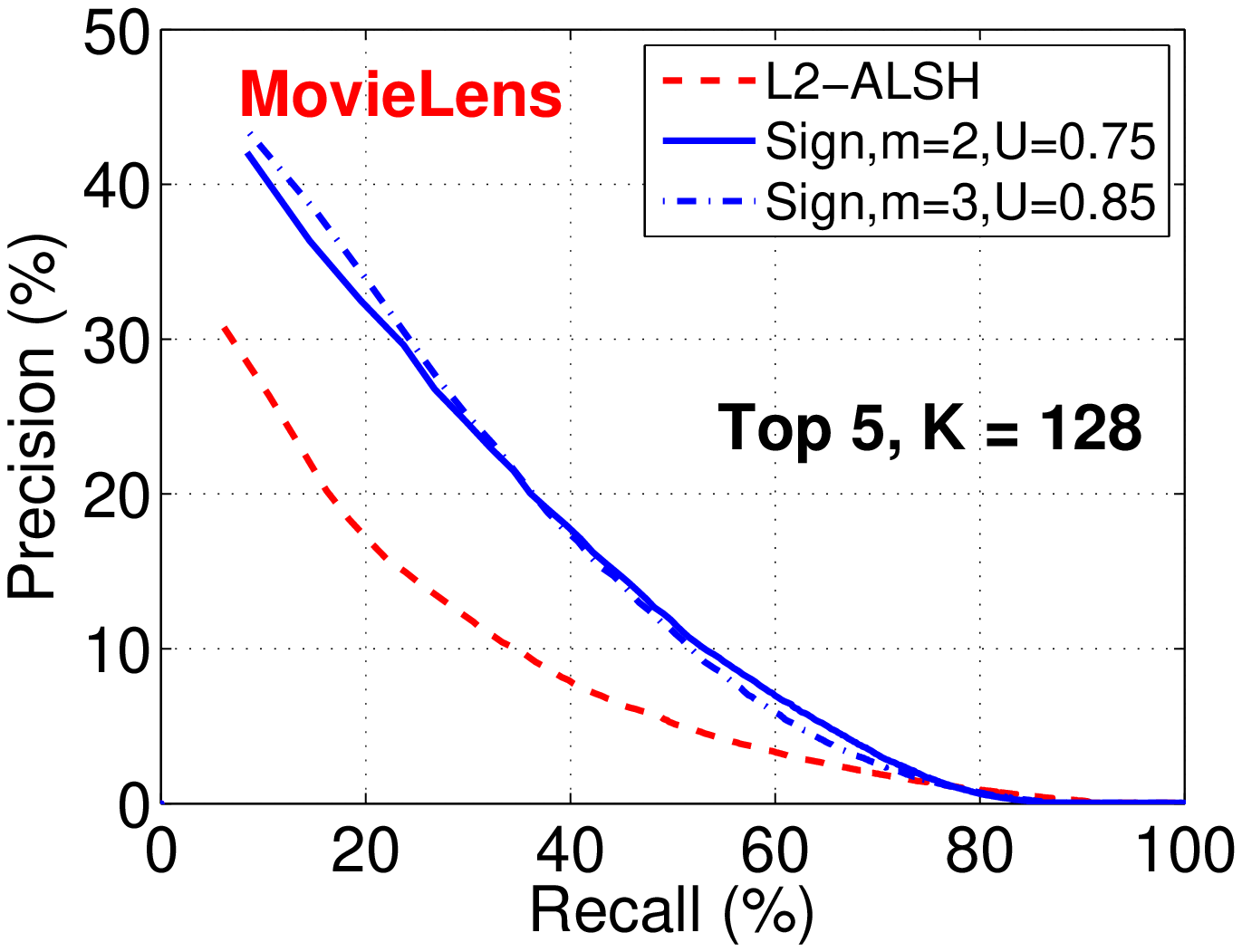}\hspace{0.13in}
\includegraphics[width=2in]{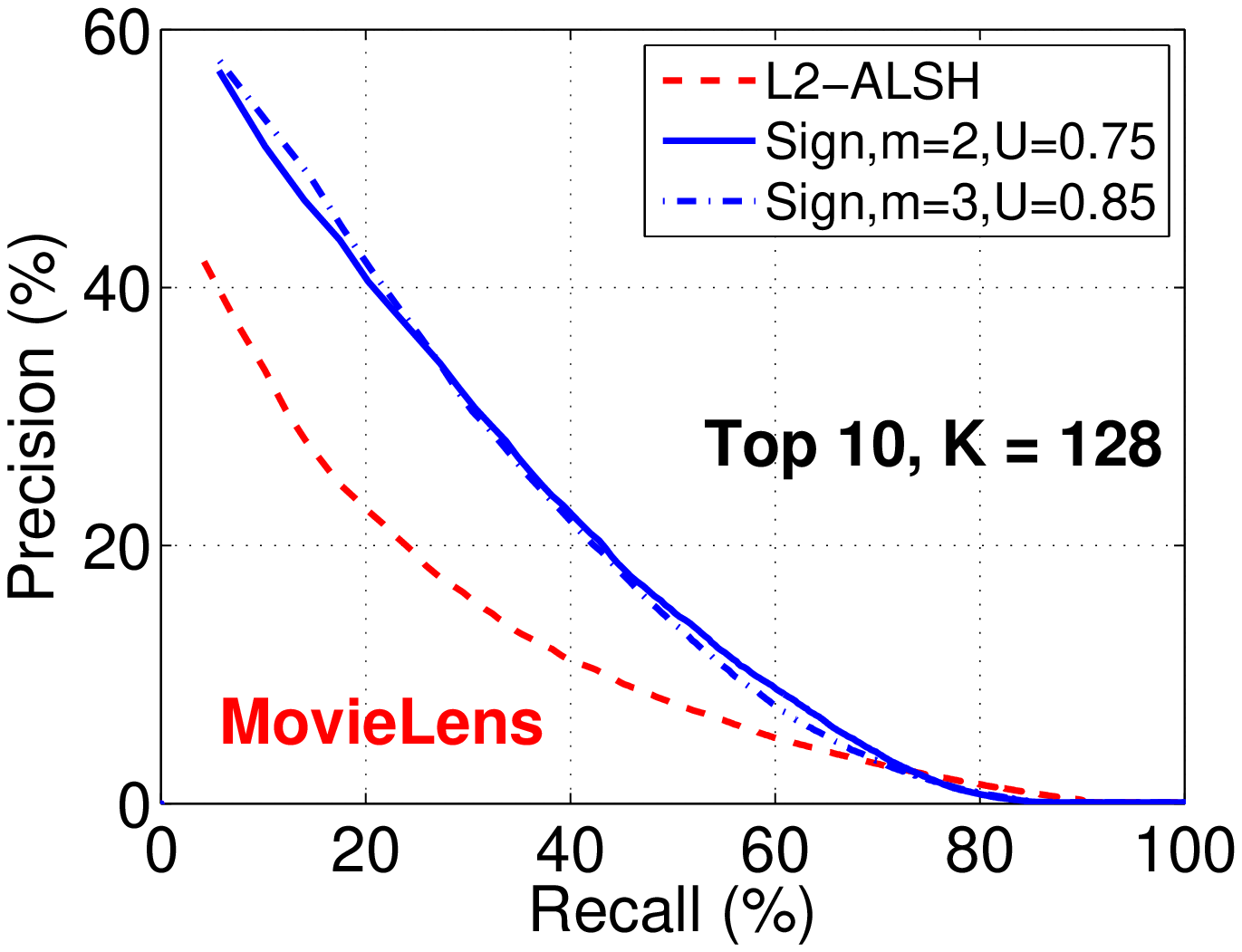}
}

\mbox{
\includegraphics[width=2in]{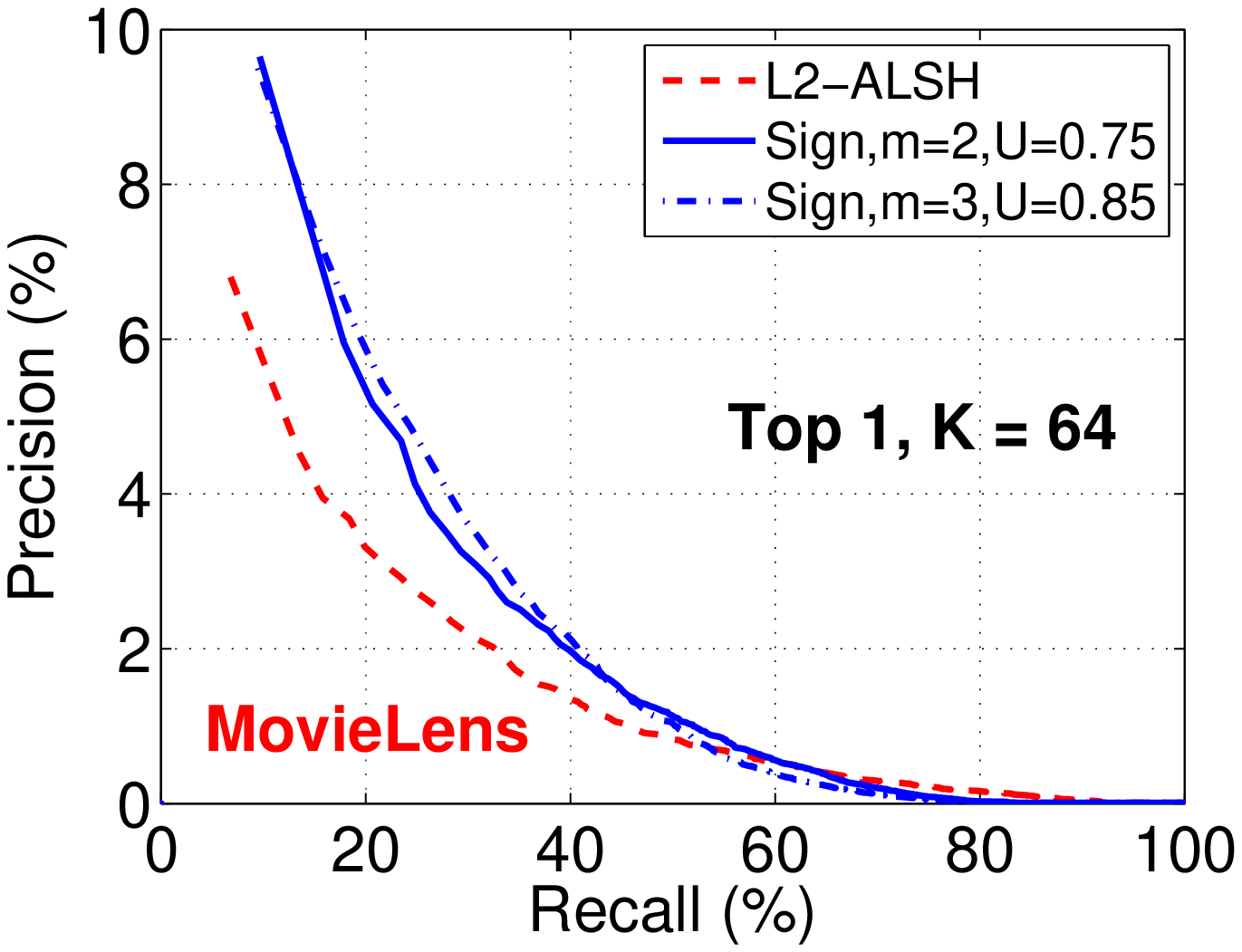}\hspace{0.13in}
\includegraphics[width=2in]{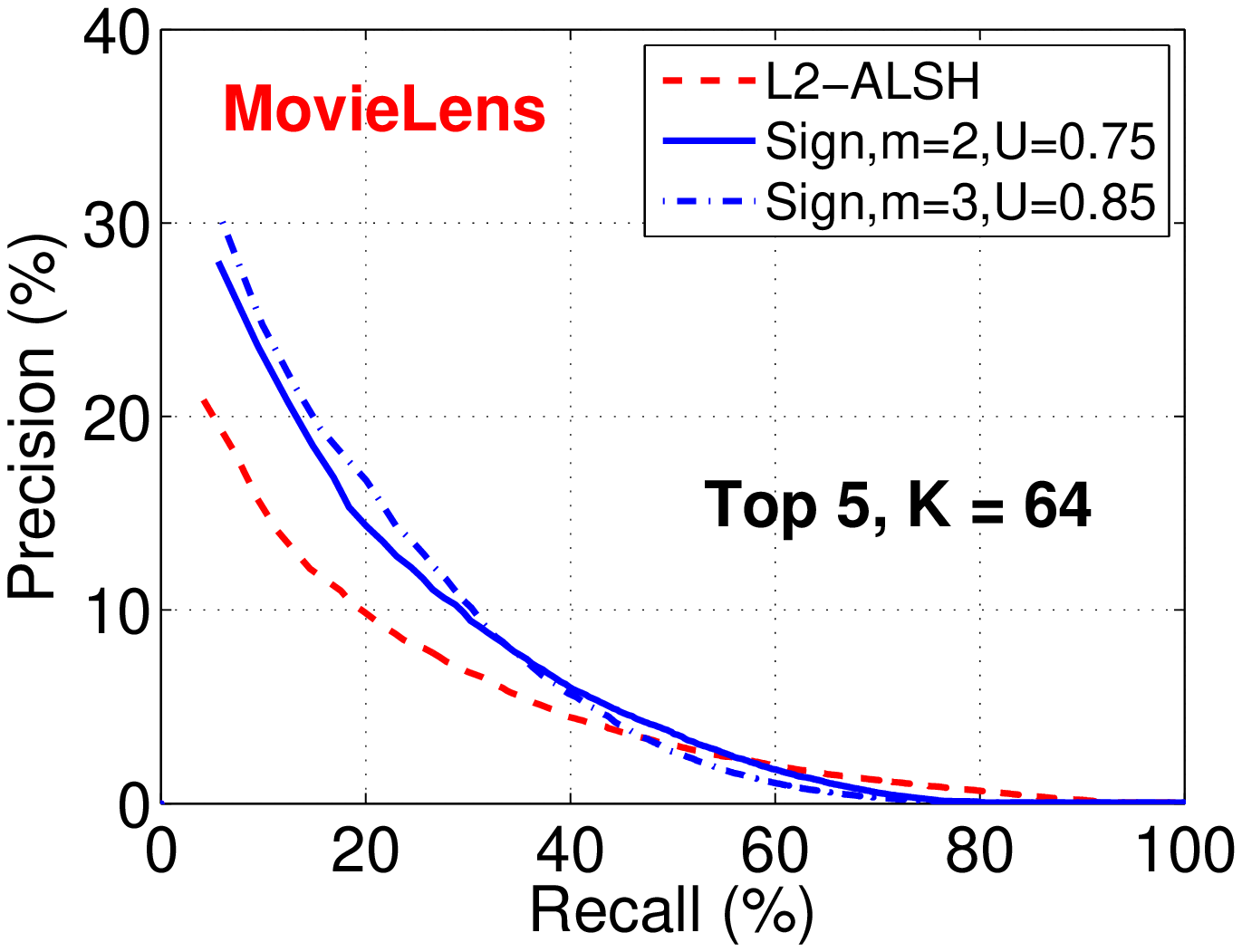}\hspace{0.13in}
\includegraphics[width=2in]{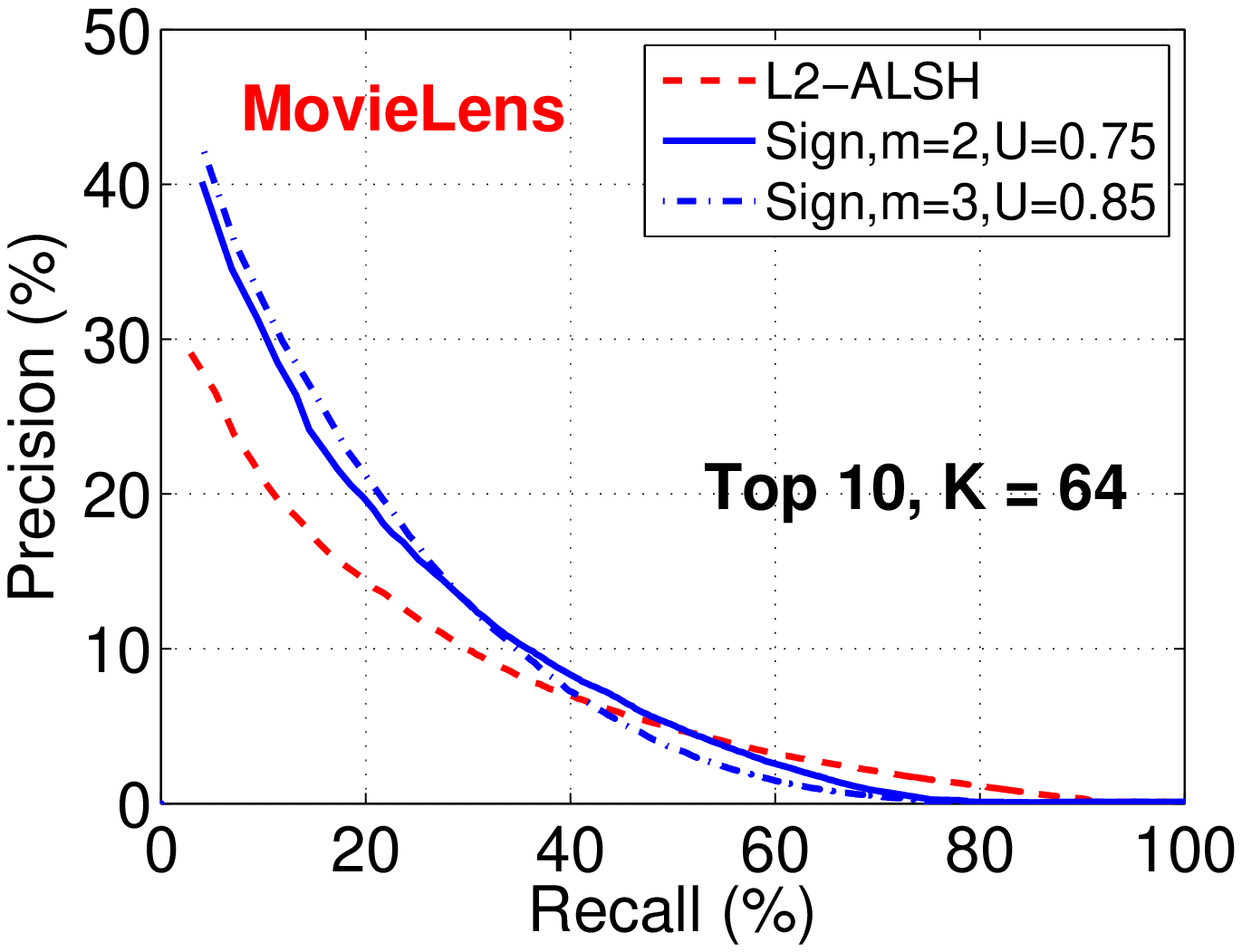}
}

\end{center}
\vspace{-0.2in}
\caption{\textbf{Movielens}. Precision-Recall curves (higher is better), of retrieving top-$T$ items, for $T=1, 5, 10$. We vary the number of hashes $K$ from 64 to 512. We compare L2-ALSH (using parameters recommended in~\cite{Report:ALSH_arXiv14}) with our proposed Sign-ALSH using two sets of parameters: $(m=2,U=0.75)$ and $(m=3,U=0.85)$.  Sign-ALSH is noticeably better. }\label{fig_MovielensRanking}\vspace{-0.15in}
\end{figure*}

Therefore the correlation or the cosine similarity between $P(Q(T(q)))$ and $Q(P(T(x)))$ is
\begin{align}
Corr= \frac{ q^Tx \times\left(\frac{U^2}{M^2}\right)}{\sqrt{\frac{m}{4} + ||T(q)||_2^{2^{m+1}}}  \sqrt{\frac{m}{4} + ||T(x)||_2^{2^{m+1}}} }
\end{align}
Note $||T(q)||_2^{2^{m+1}}, ||T(x)||_2^{2^{m+1}} \le U < 1$, therefore both $||T(q)||_2^{2^{m+1}}$ and $||T(x)||_2^{2^{m+1}}$ converge to zero at a tower rate and we get approximate monotonicity of correlation with the inner products. We can apply sign random projections to hash $P(Q(T(q)))$ and $Q(P(T(q)))$.

\begin{figure*}[ht]
\vspace{-0.15in}
\begin{center}
\mbox{
\includegraphics[width=2in]{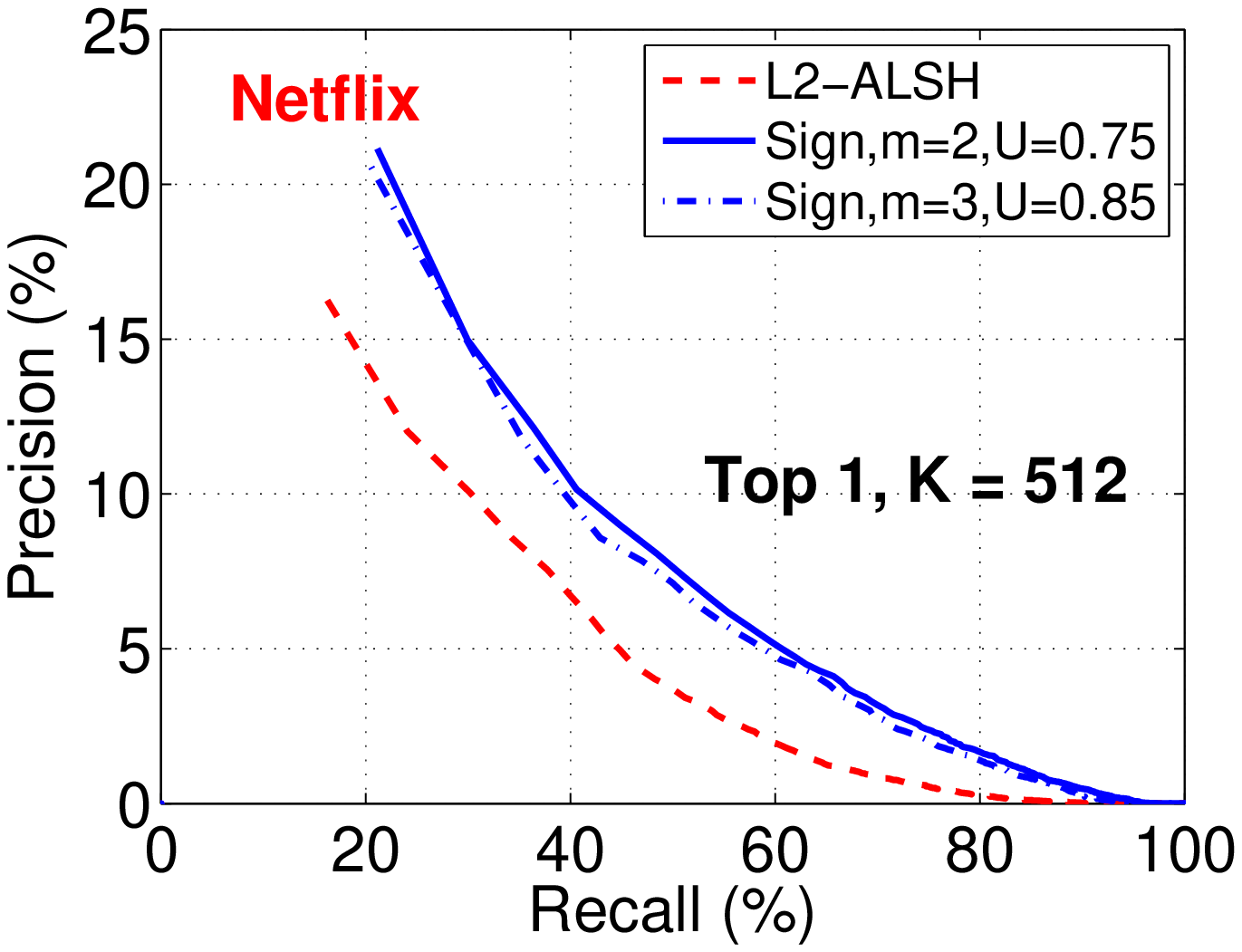}\hspace{0.13in}
\includegraphics[width=2in]{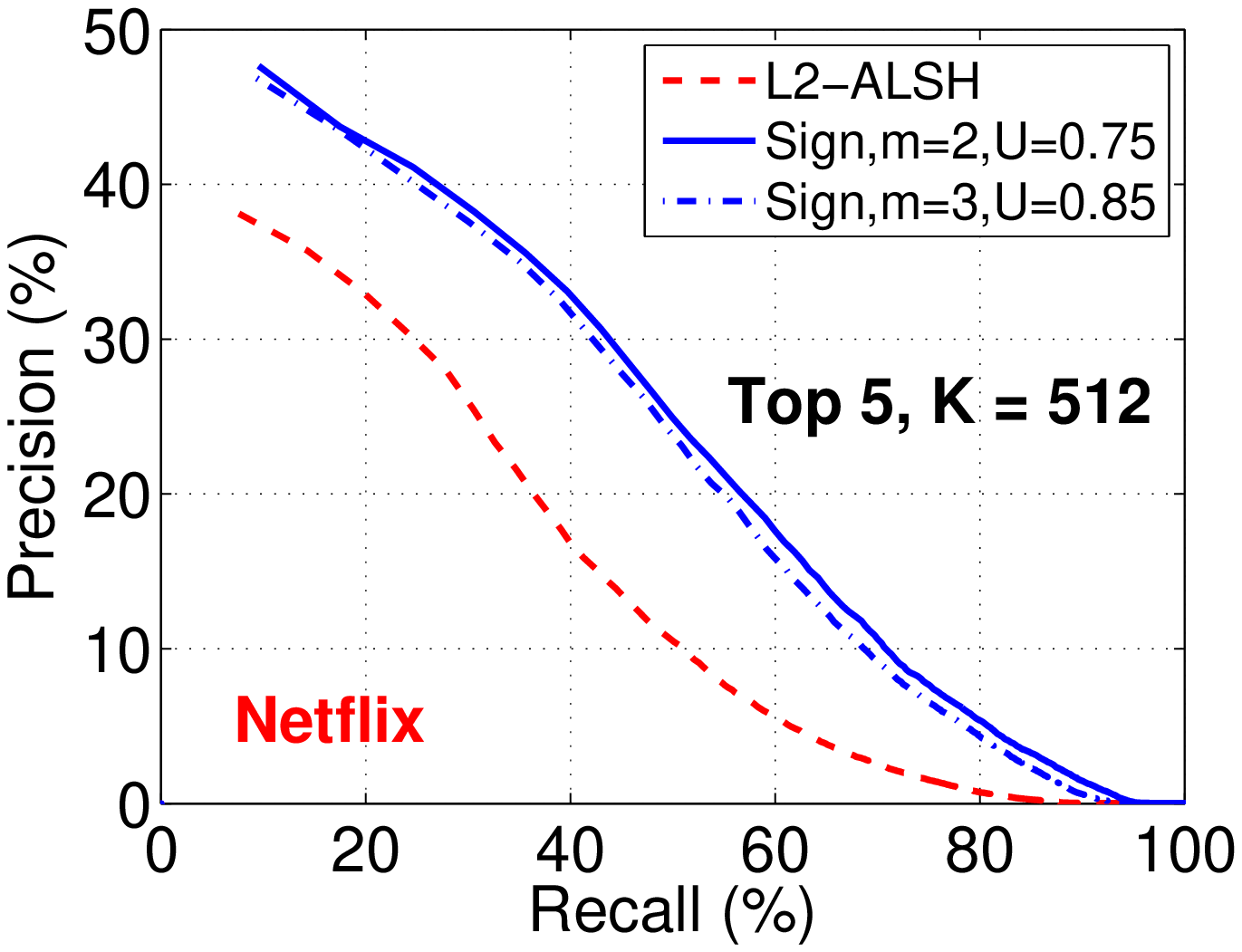}\hspace{0.13in}
\includegraphics[width=2in]{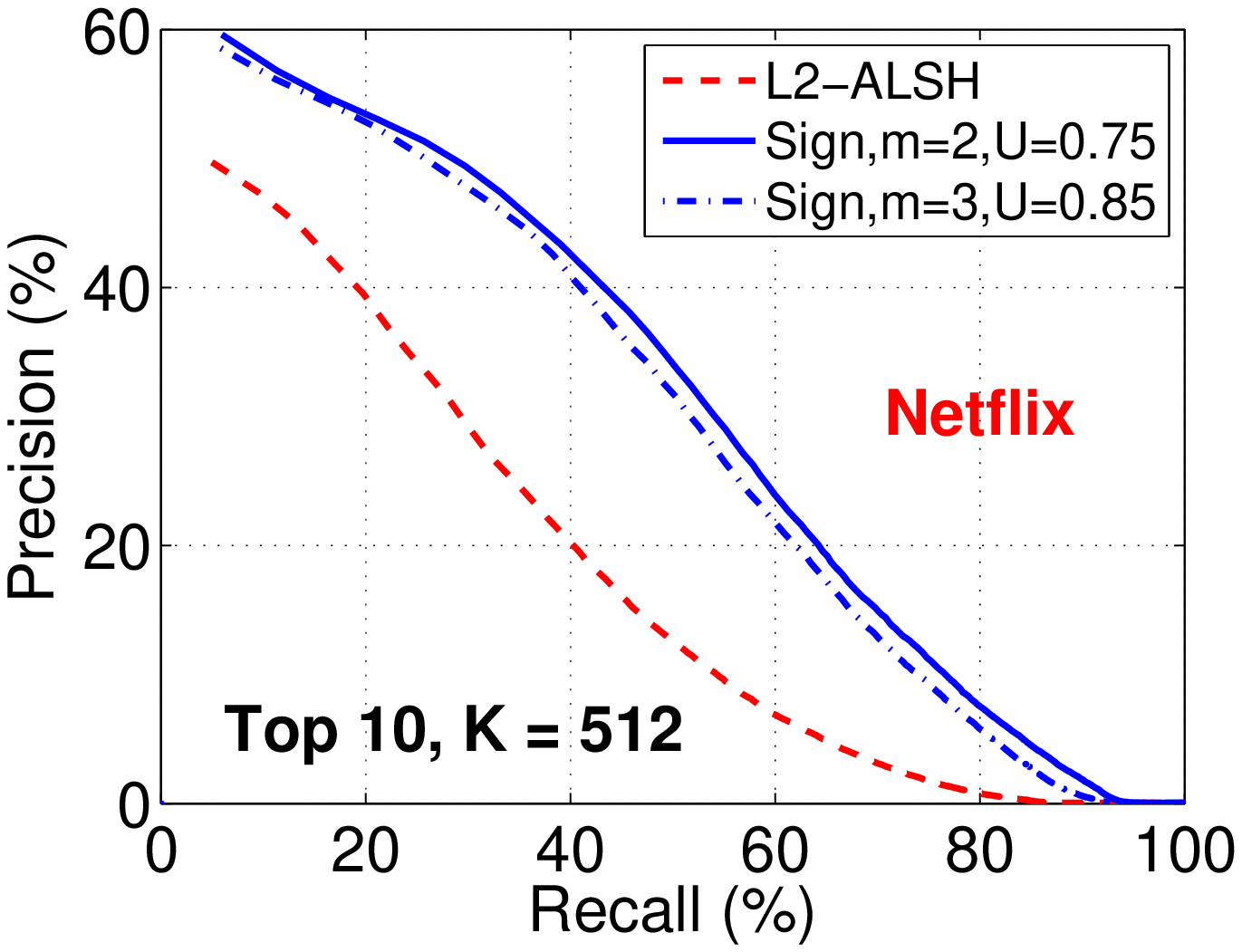}
}

\mbox{
\includegraphics[width=2in]{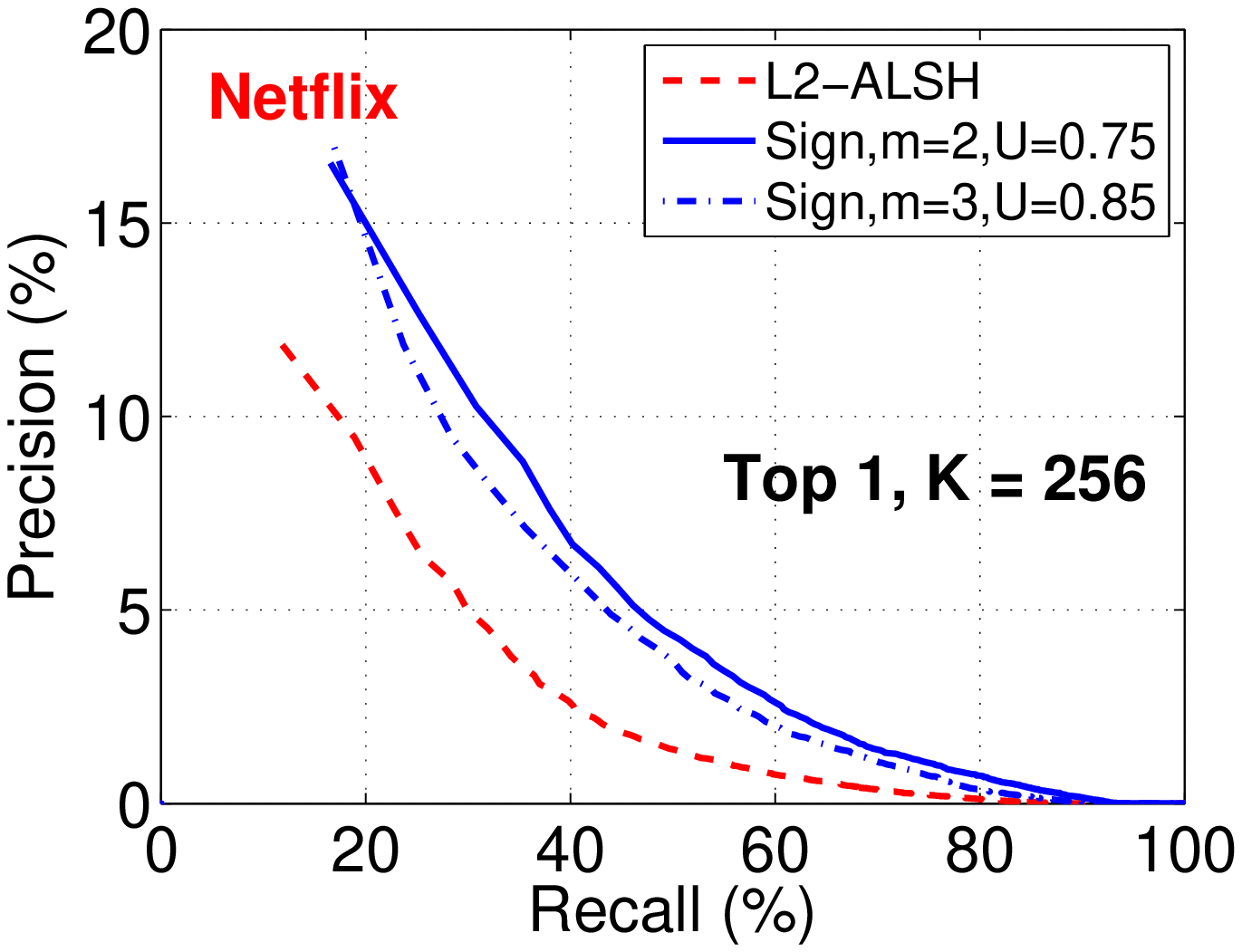}\hspace{0.13in}
\includegraphics[width=2in]{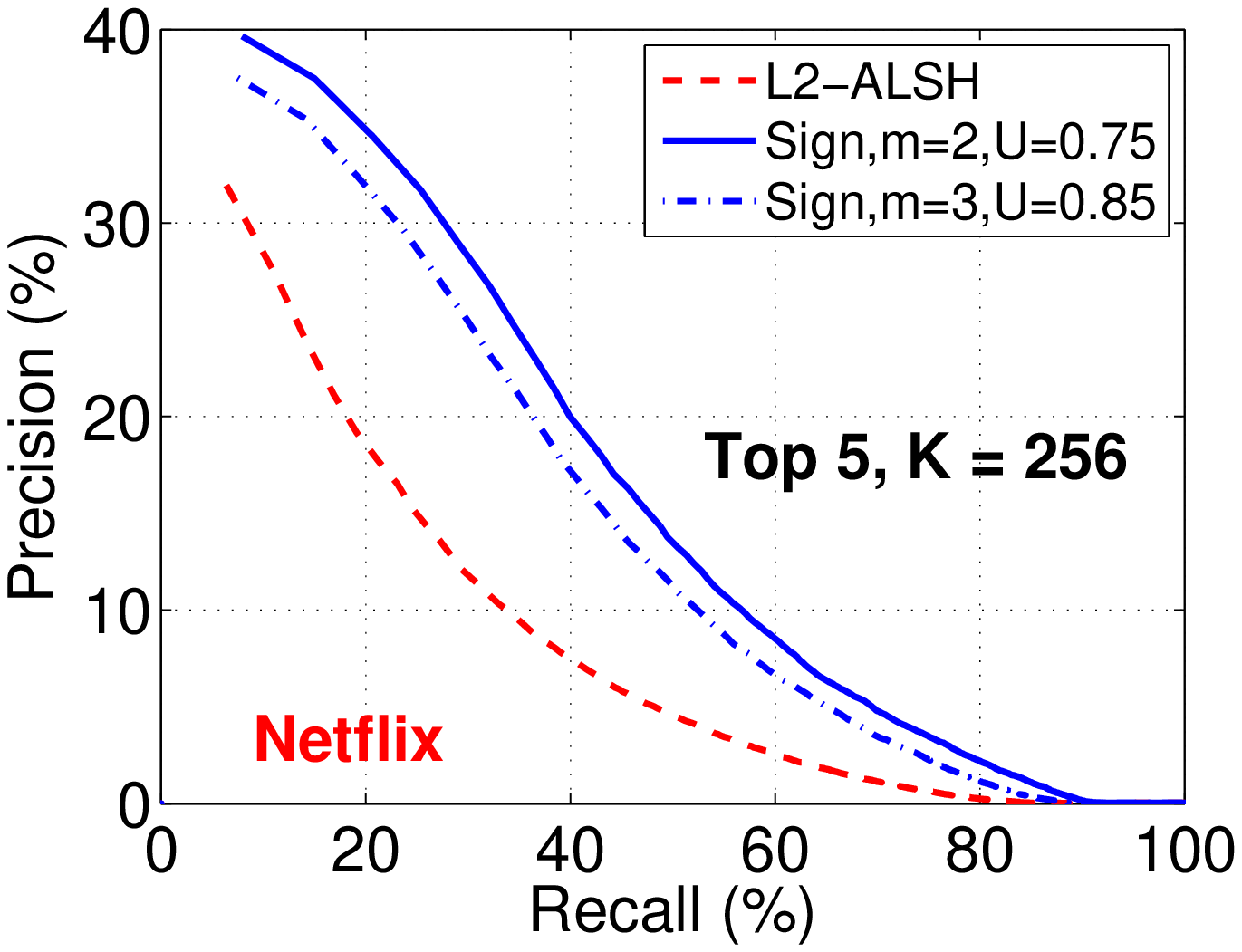}\hspace{0.13in}
\includegraphics[width=2in]{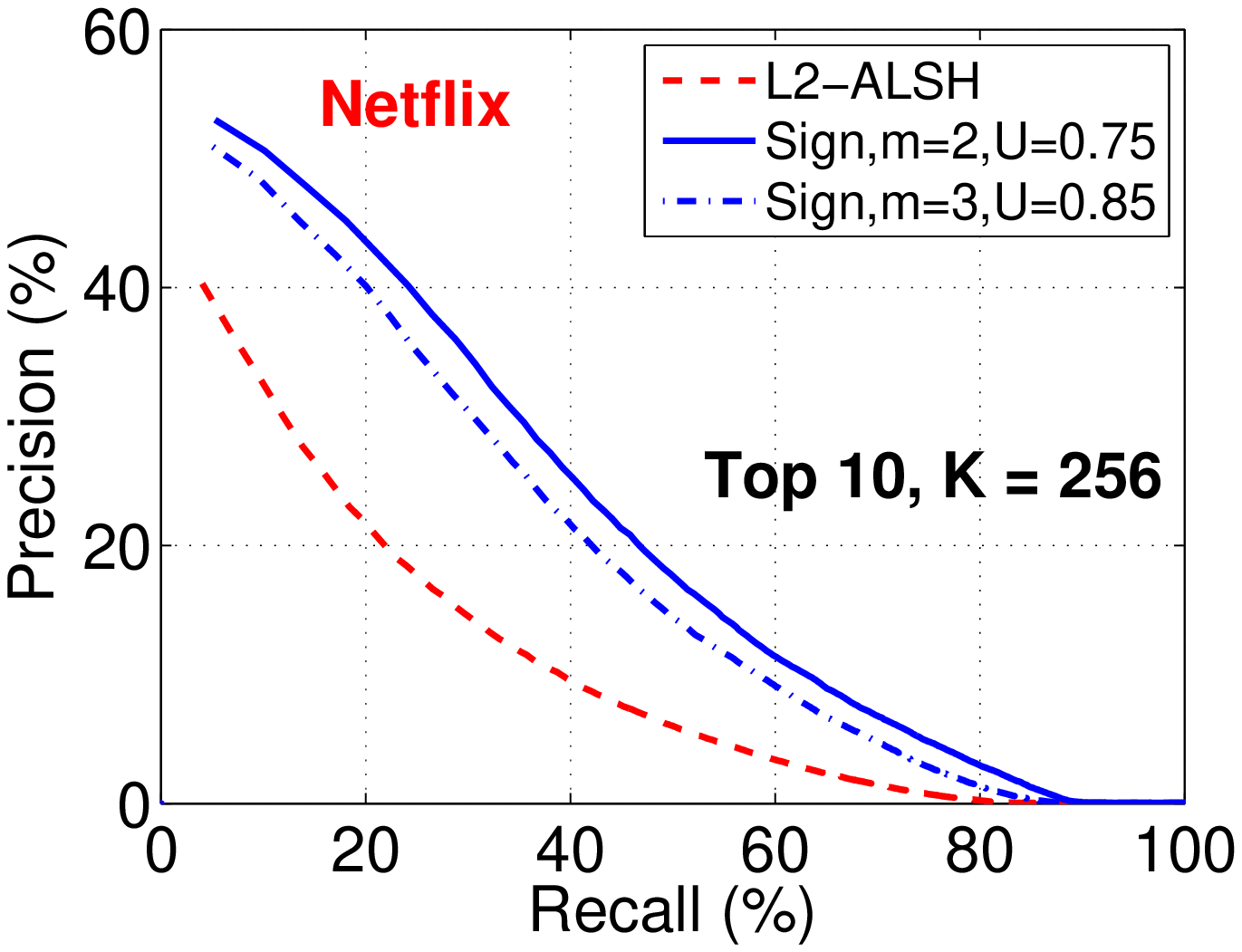}
}

\mbox{
\includegraphics[width=2in]{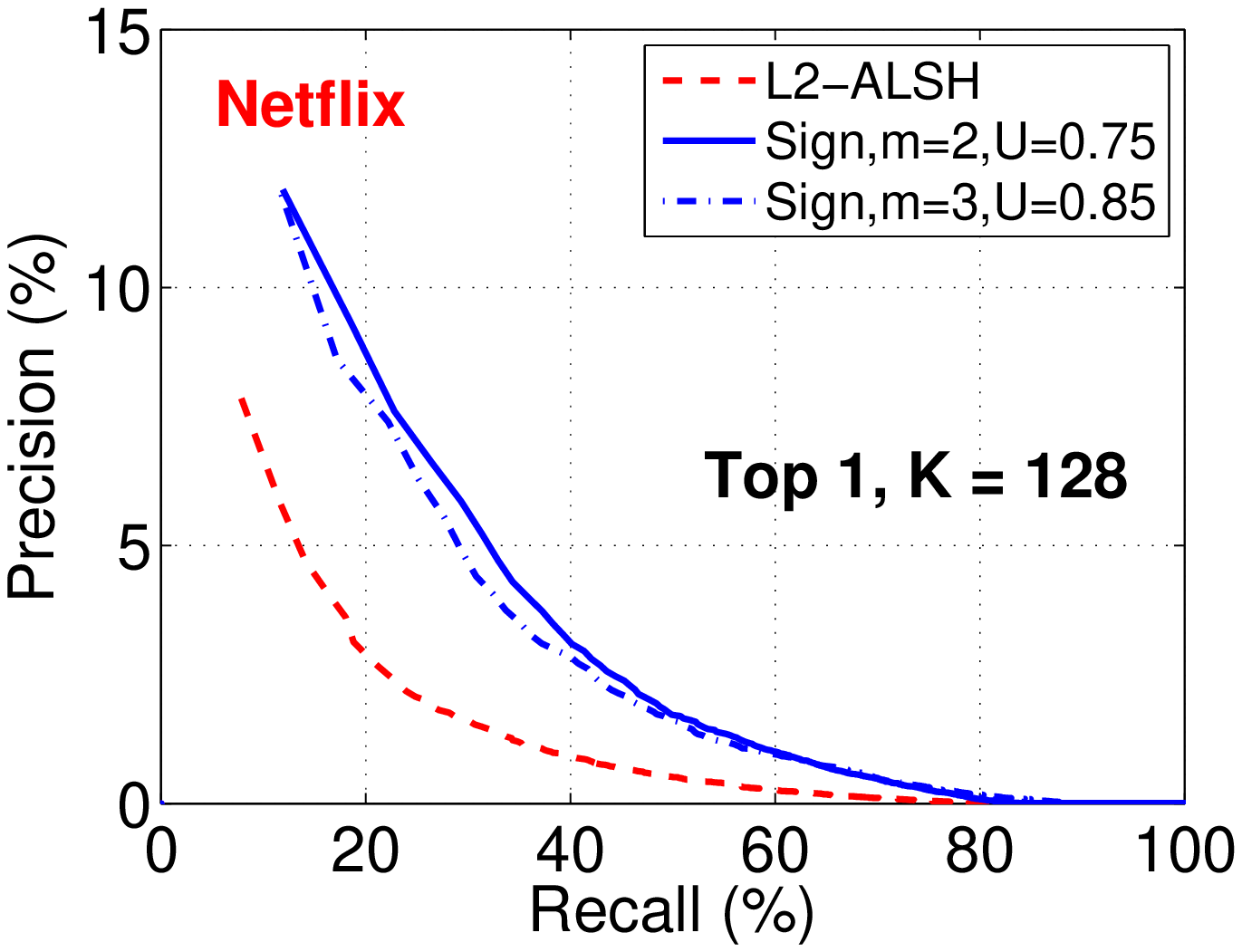}\hspace{0.13in}
\includegraphics[width=2in]{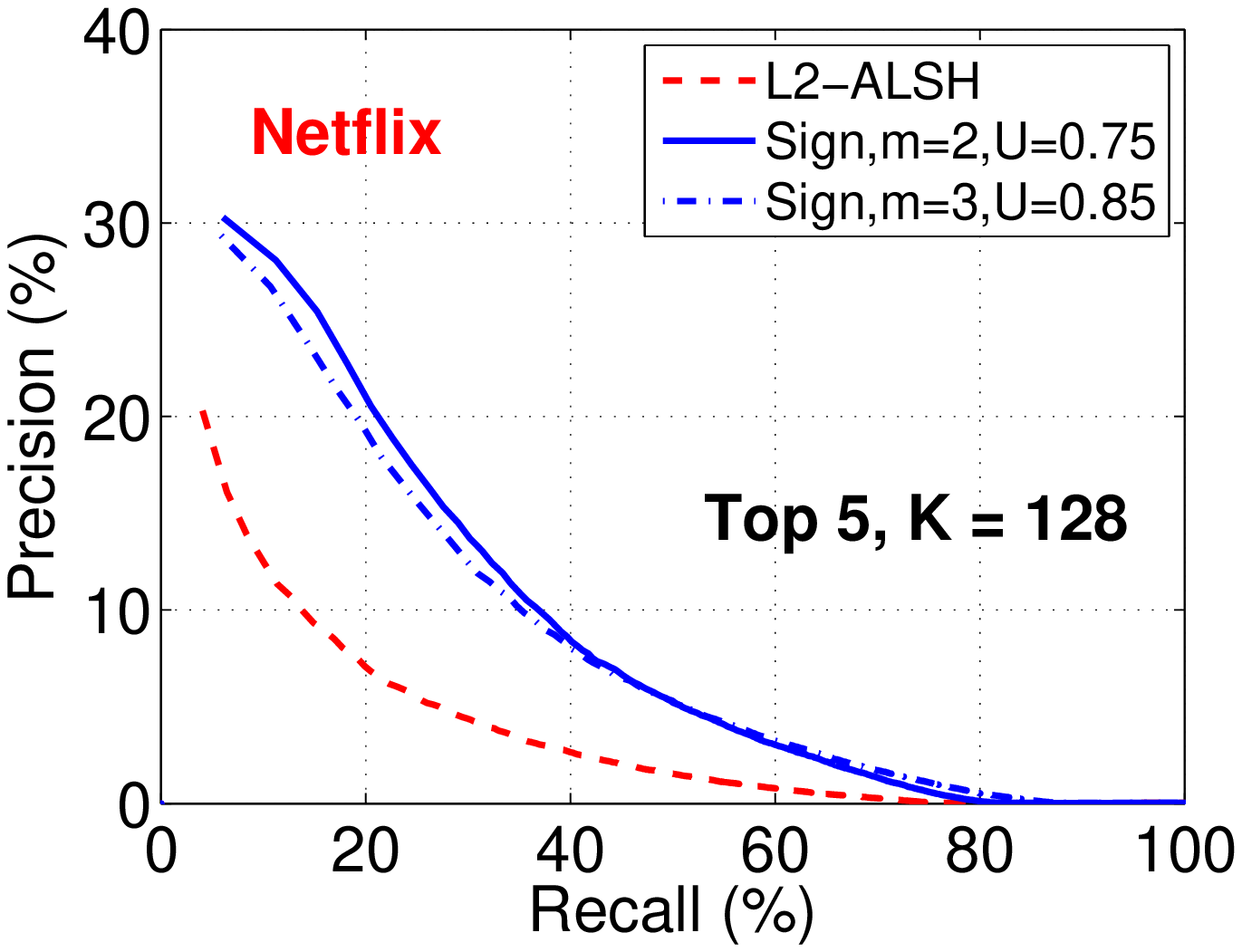}\hspace{0.13in}
\includegraphics[width=2in]{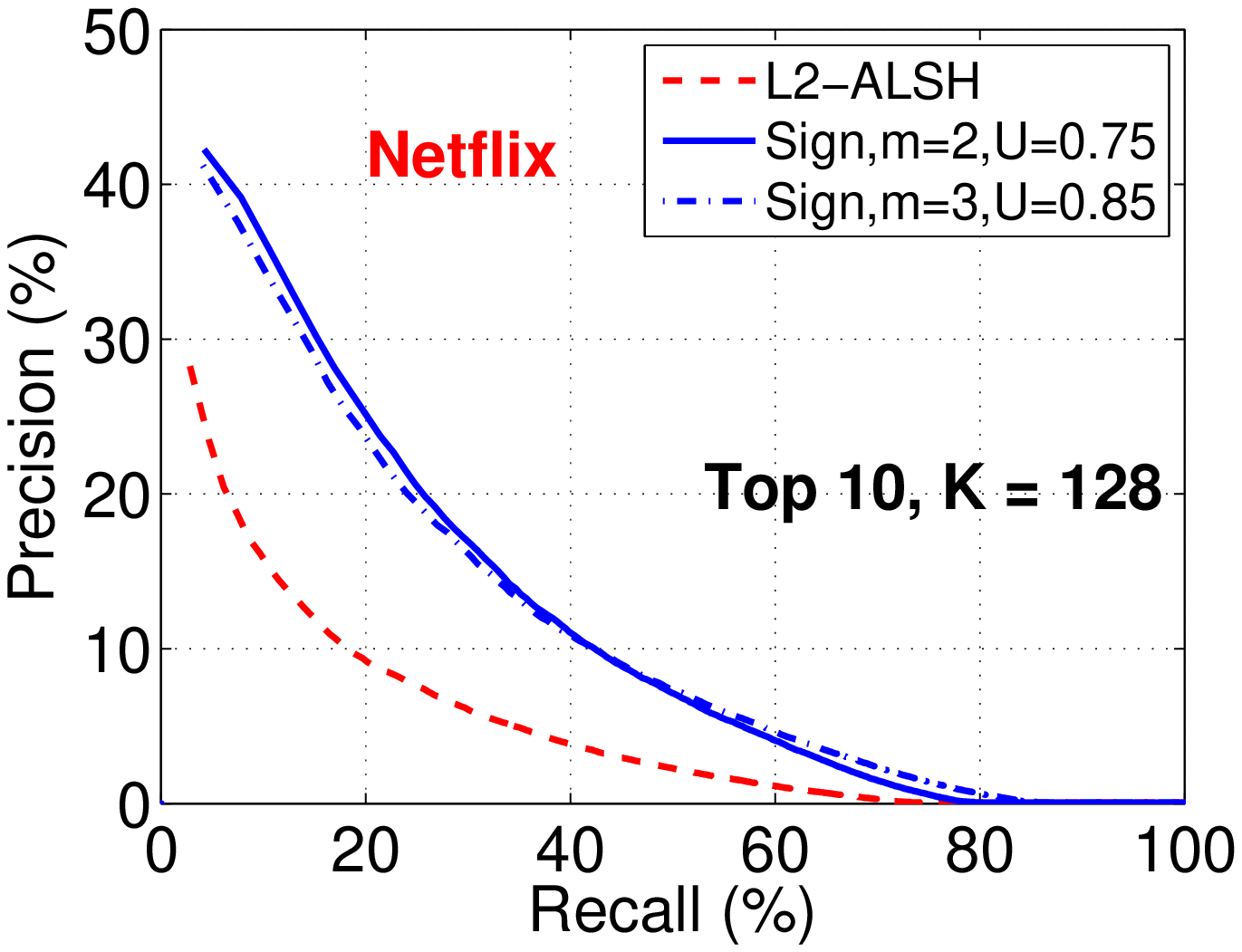}
}

\mbox{
\includegraphics[width=2in]{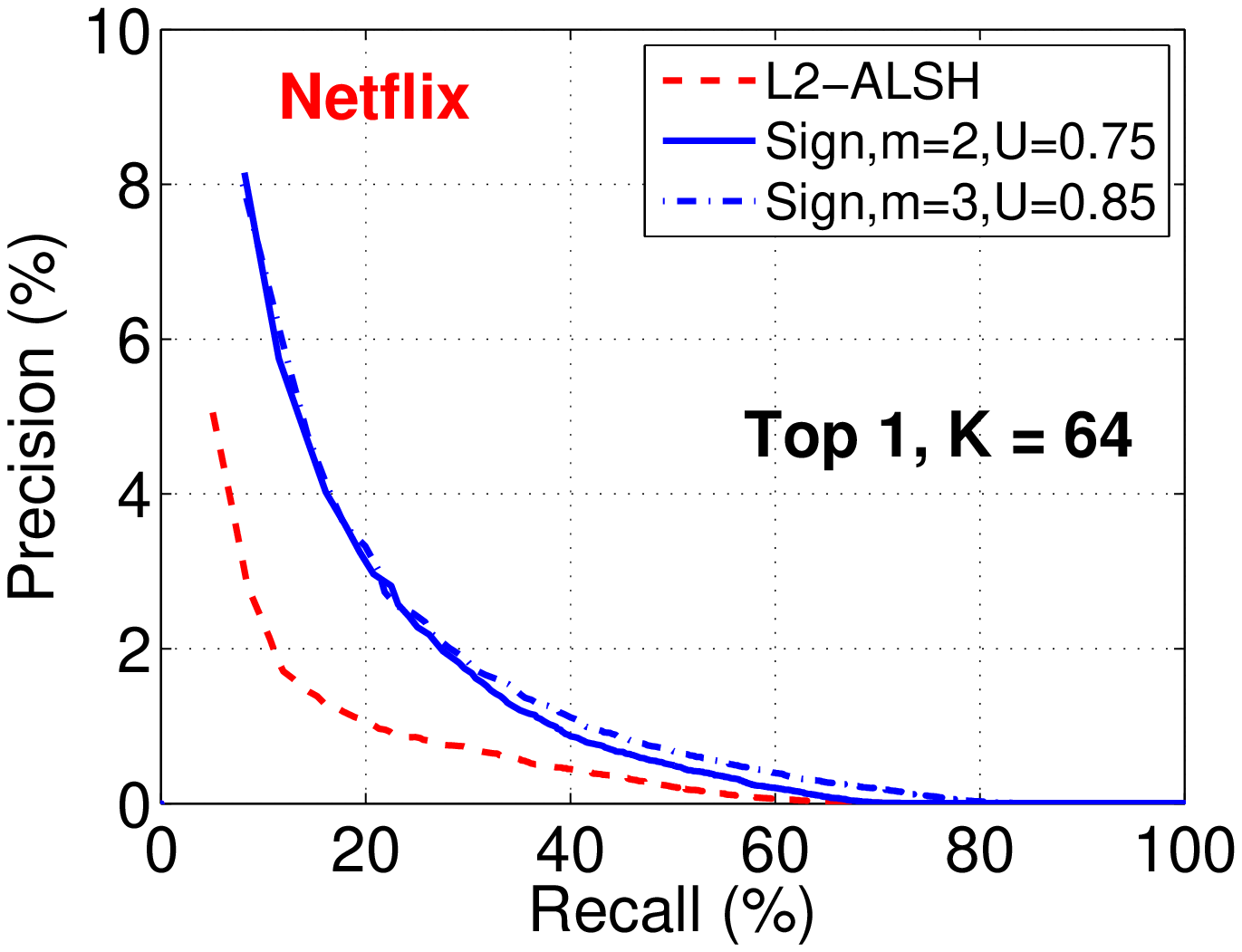}\hspace{0.13in}
\includegraphics[width=2in]{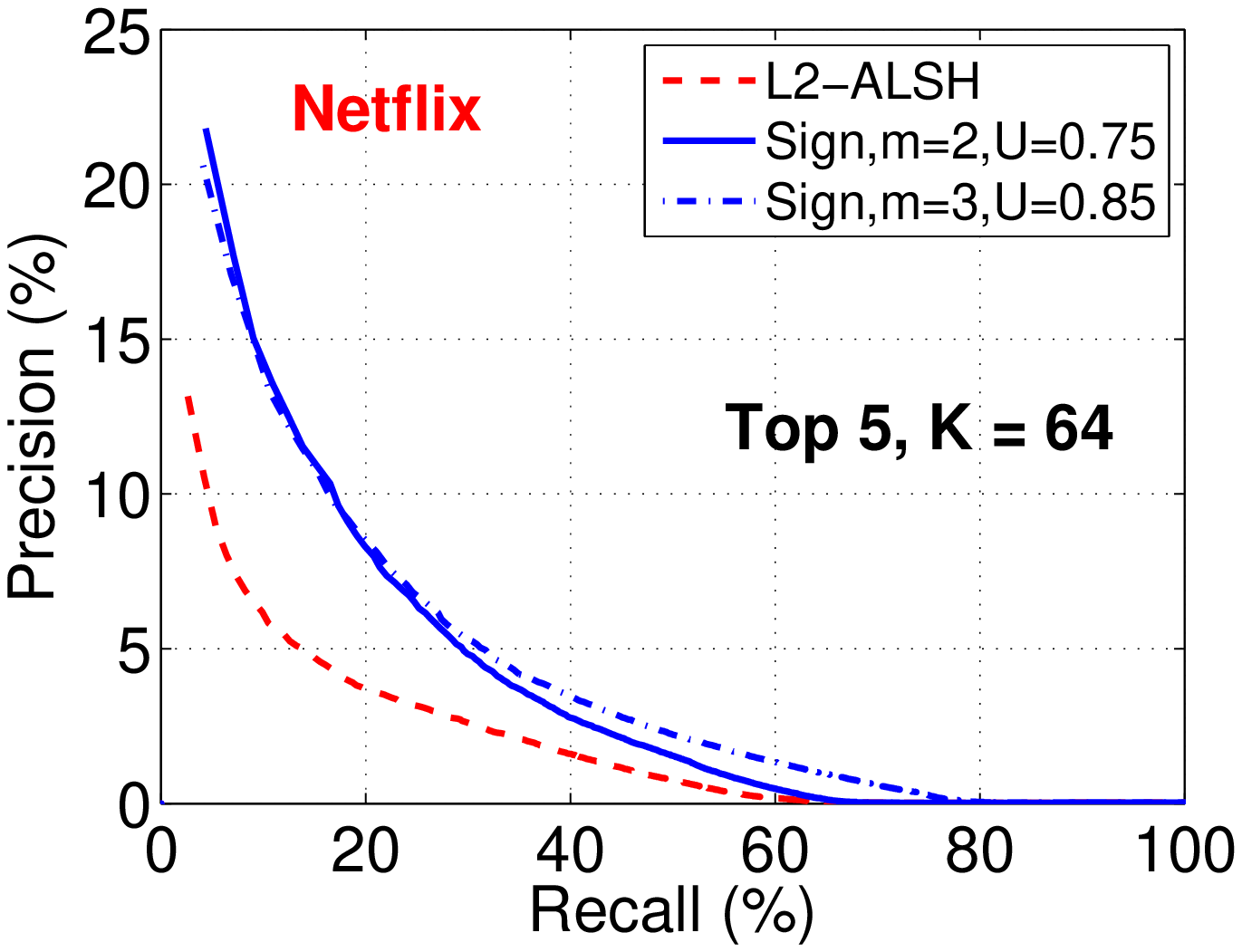}\hspace{0.13in}
\includegraphics[width=2in]{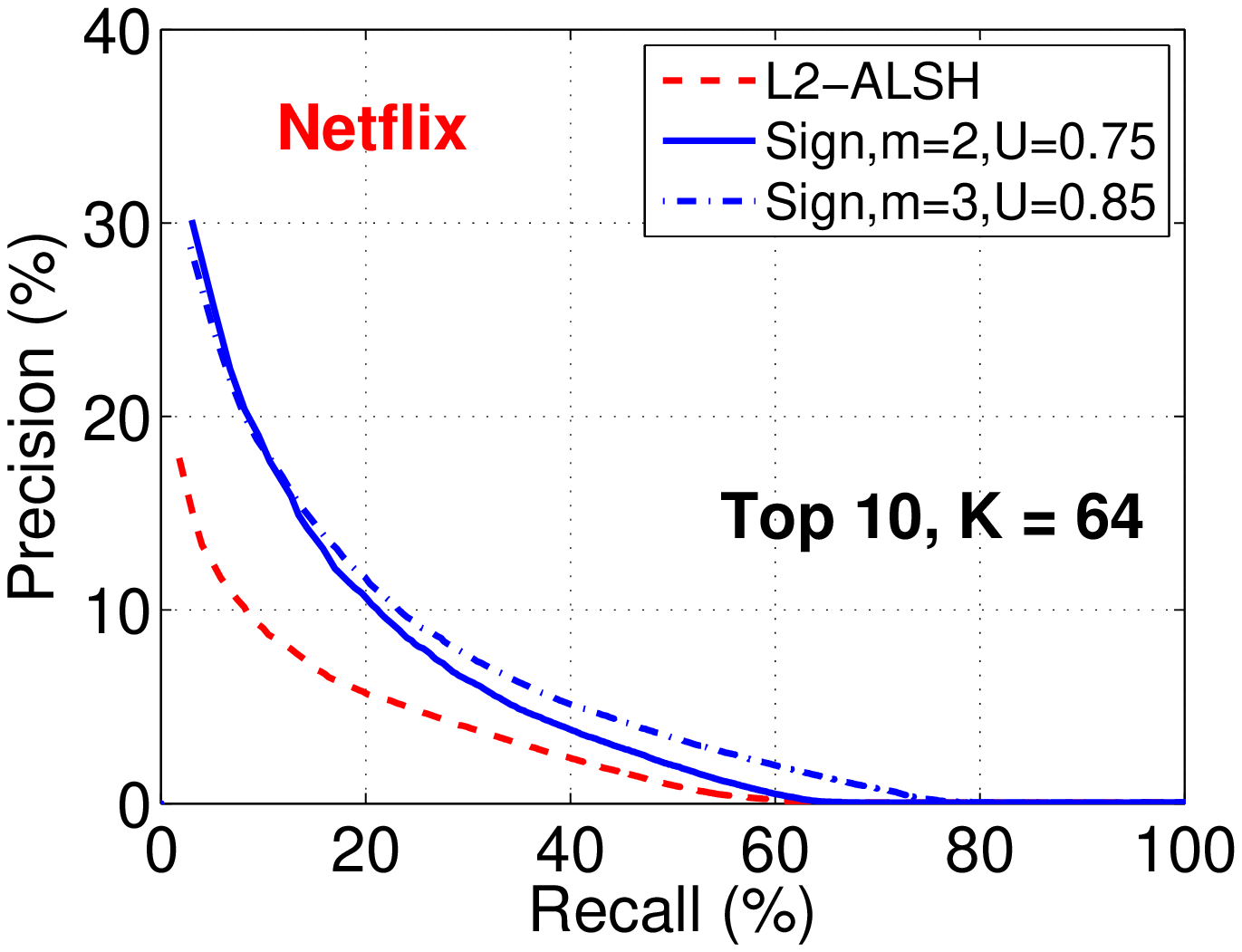}
}
\end{center}
\vspace{-0.2in}
\caption{\textbf{Netflix}. Precision-Recall curves (higher is better), of retrieving top-$T$ items, for $T=1, 5, 10$. We vary the number of hashes $K$ from 64 to 512. We compare L2-ALSH (using parameters recommended in~\cite{Report:ALSH_arXiv14}) with our proposed Sign-ALSH using two sets of parameters: $(m=2,U=0.75)$ and $(m=3,U=0.85)$. Sign-ALSH is noticeably better.   }\label{fig_NetflixRanking}\vspace{-0.15in}
\end{figure*}

Using the fact $0 \le ||T(q)||_2^{2^{m+1}} \le U$ and $ 0\le ||T(x)||_2^{2^{m+1}} \le U$, it is not difficult to  get $p_1$ and $p_2$ for Sign-ALSH, without any conditions on any norms. Simplifying the expression, we get the following value of optimal $\rho_u$ (u for unrestricted).
\begin{align}
\label{eq:optrho_u}
\rho_u^* &= \min_{U,m,} \frac{\log{\bigg(1-\frac{1}{\pi}\cos^{-1}\bigg(\frac{ S_0 \times\left(\frac{U^2}{M^2}\right)}{\frac{m}{4} + U^{2^{m+1}}}\bigg)\bigg)}}{\log{\bigg(1-\frac{1}{\pi}\cos^{-1} \bigg(\frac{ cS_0 \times\left(\frac{4U^2}{M^2}\right)}{m}\bigg)\bigg)}}\\
\nonumber s.t. \hspace{0.1in} U^{2^{m+1}} &<  \frac{m(1-c)}{4c}, \hspace{0.05in}  m \in \mathbb{N}^{+}, \mbox{ and } 0 < U < 1.
\end{align}
 With this value of $\rho_u^*$, we can state our main theorem.
\begin{theorem}
\label{theo:main}  For the problem of $c$-approximate MIPS in a bounded space, one can construct a data structure having $O(n^{\rho_u^*} \log{n})$ query time and space $O(n^{1+\rho_u^*})$, where $\rho_u^* < 1$  is the solution to constraint optimization (\ref{eq:optrho_u}).
\end{theorem}
Note, for all $c < 1$, we always have $\rho_u^* < 1$ because the constraint
$U^{2^{m+1}} <  \frac{m(1-c)}{4c}$ is always true for big enough $m$. The only assumption for efficiently solving MIPS that we need is that the space is bounded, which is always satisfied for any finite dataset. $\rho_u^*$ depends on $M$, the radius of the space, which is expected.

\newpage

\vspace{-0.15in}
\section{Ranking Evaluations}
\label{sec:evaluations}
\vspace{-0.1in}

\begin{figure*}[ht]
\vspace{-0.2in}
\begin{center}
\mbox{
\includegraphics[width=2.3in]{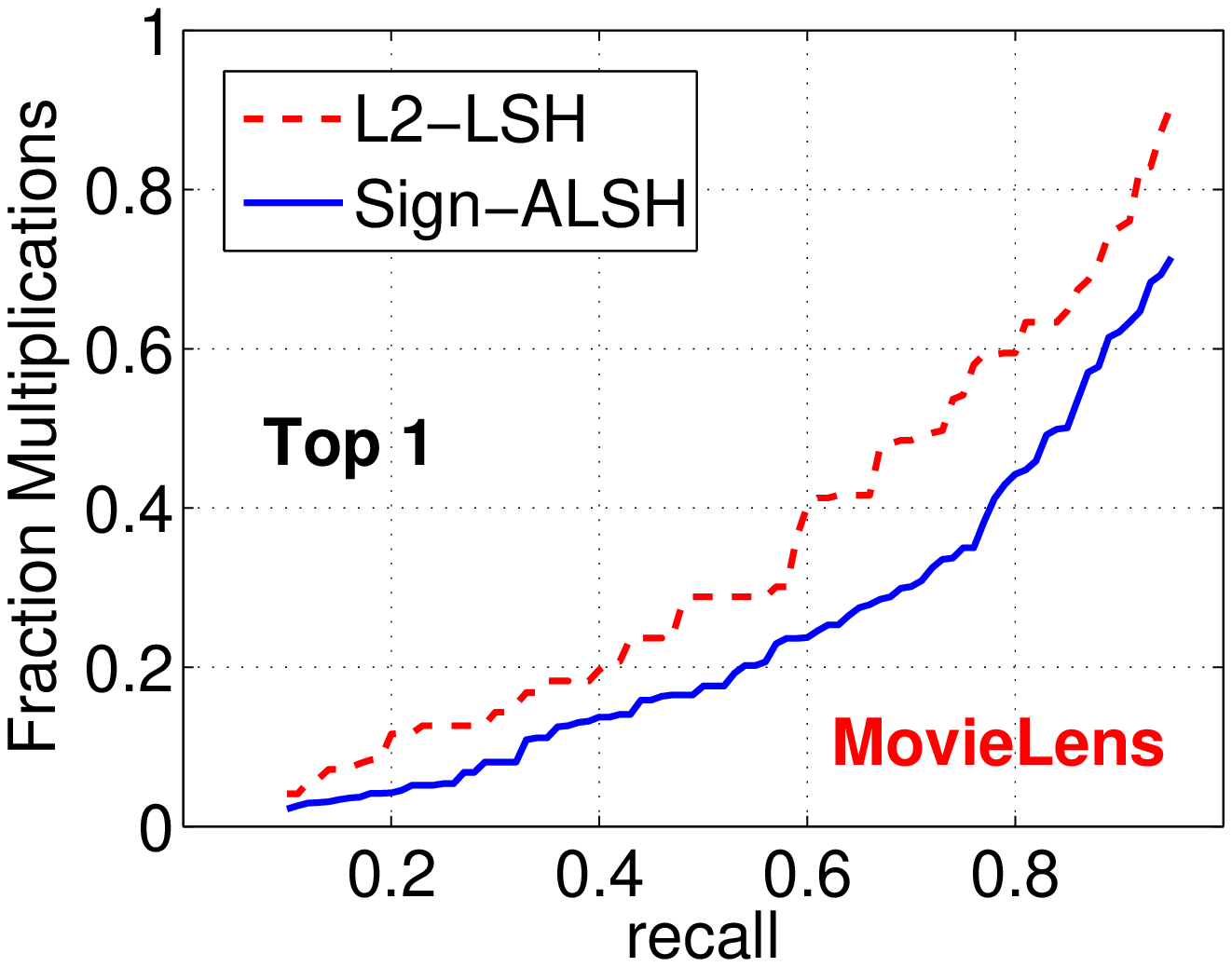}\hspace{-0.15in}
\includegraphics[width=2.3in]{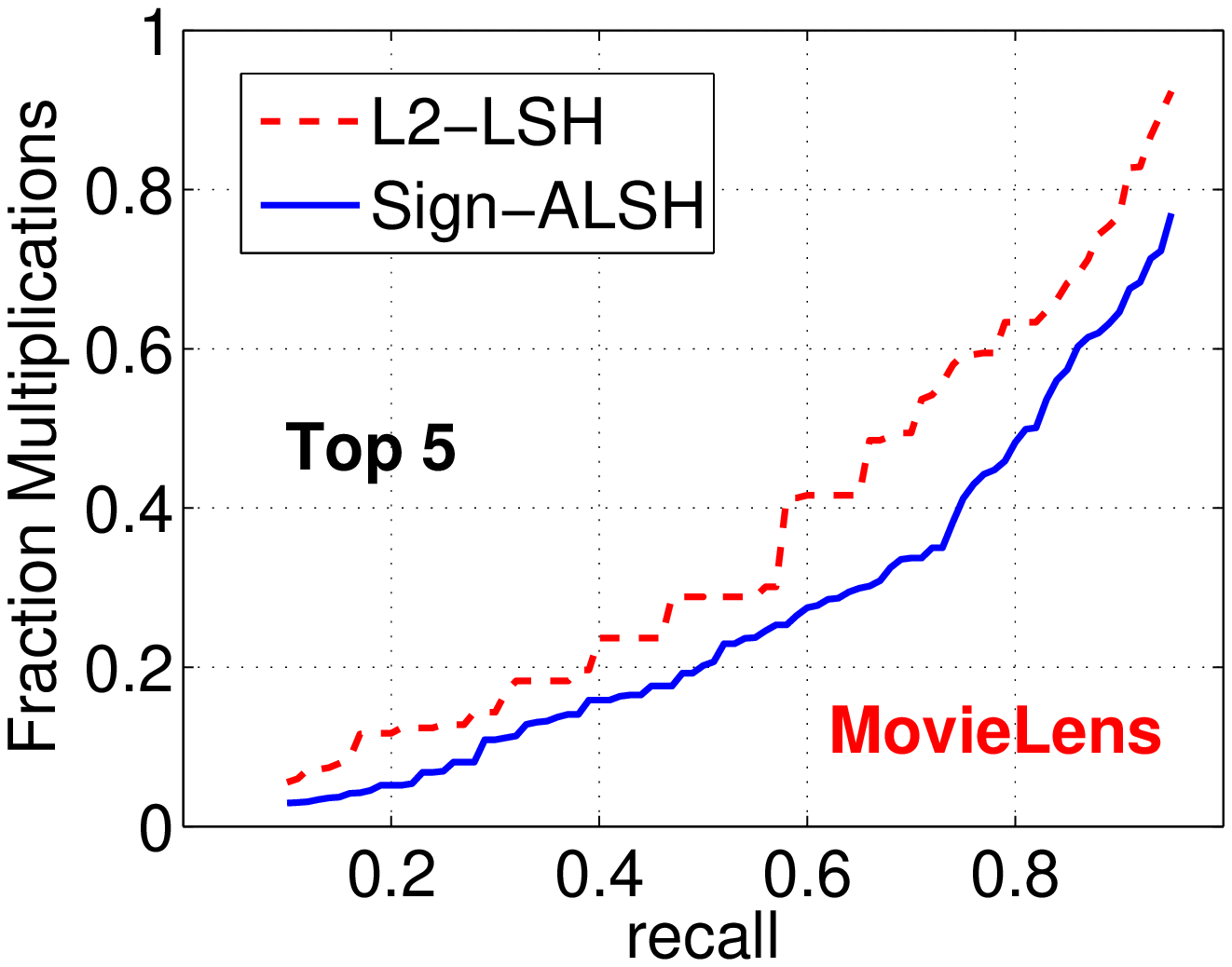}\hspace{-0.15in}
\includegraphics[width=2.3in]{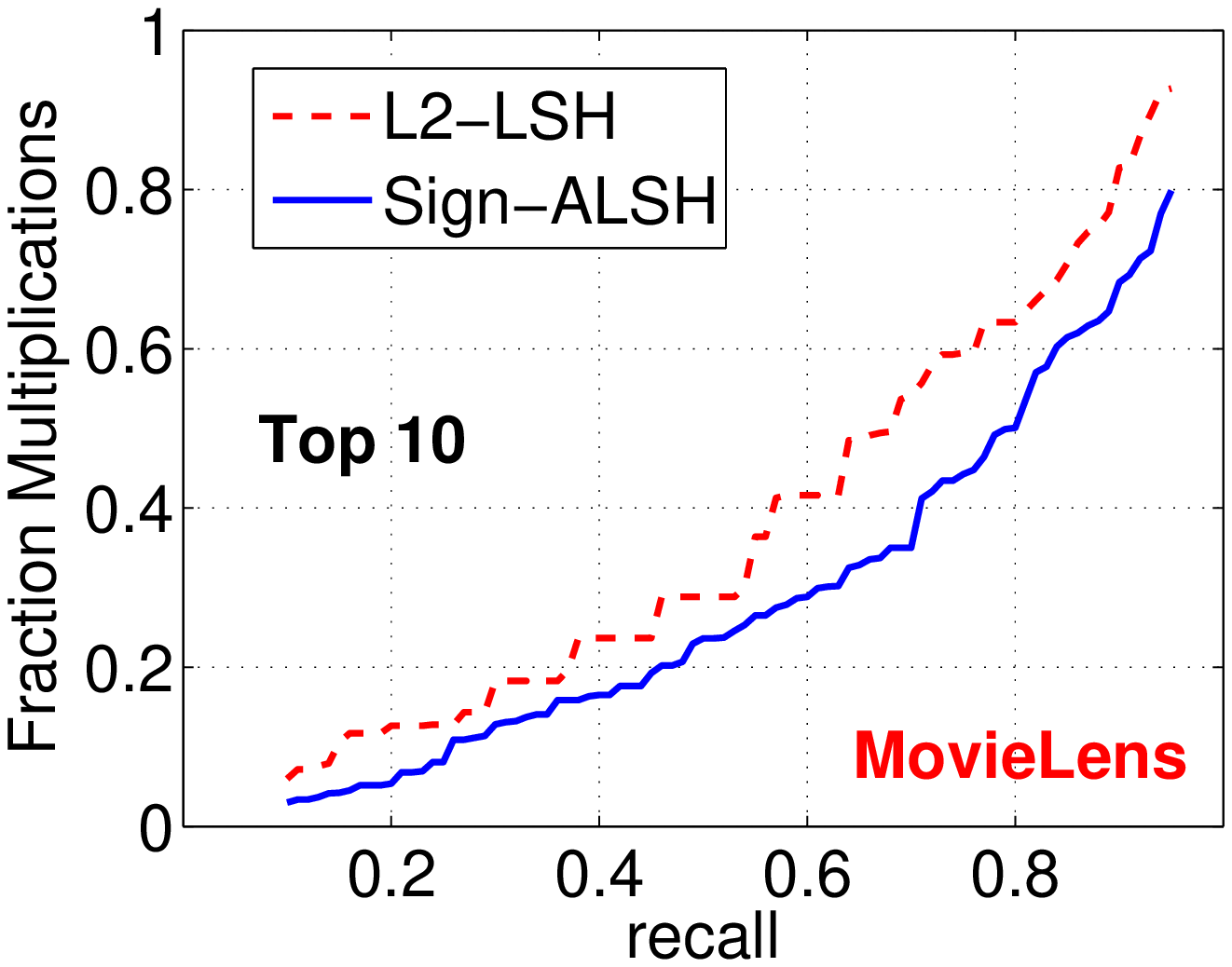}
}
\end{center}
\vspace{-0.2in}
\caption{\textbf{MovieLens}. Recall-FIP (Fractions of Inner Products) curves for top-1, top-5, and top-10, for Sign-ALSH with L2-ALSH.   We used the recommended parameters for L2-ALSH~\cite{Report:ALSH_arXiv14}. For Sign-ALSH, we used $m=2$ and $U=0.75$. }\label{fig_MovieLensLSH}\vspace{-0.15in}
\end{figure*}

In~\cite{Report:ALSH_arXiv14}, the L2-ALSH scheme was shown to outperform the LSH for L2 distance in retrieving maximum inner products.  Since our proposal is an improvement over L2-ALSH, we focus on comparisons  with L2-ALSH. In this section, we compare L2-ALSH with Sign-ALSH based on ranking.

\vspace{-0.05in}
\subsection{Datasets}
\vspace{-0.05in}

We use the two popular collaborative filtering datasets {\bf MovieLens 10M} and {\bf Netflix}, for the task of item recommendations. These are also the same datasets used in~\cite{Report:ALSH_arXiv14}. Each dataset is a sparse \textbf{user-item matrix} $R$, where $R(i,j)$ indicates the rating of user $i$ for movie $j$. For getting the latent feature vectors from user item matrix, we follow the methodology of ~\cite{Report:ALSH_arXiv14}. They use PureSVD procedure described in~\cite{Proc:Cremonesi_RecSys} to generate user and item latent vectors, which involves computing the SVD of $R$
$$R = W\Sigma V^T$$
where $W$ is $n_{users} \times f$ matrix and $V$ is $n_{item} \times f$ matrix for some chosen rank $f$ also known as latent dimension.

After the SVD step, the rows of matrix $U = W\Sigma$ are treated as the user characteristic vectors while rows of matrix $V$ correspond to the item characteristic vectors. This simple procedure has been shown to outperform other popular recommendation algorithms for the task of top item recommendations in~\cite{Proc:Cremonesi_RecSys}, on these two datasets. We use the same choices for the latent dimension  $f$, i.e., $f=150$ for Movielens and $f=300$ for Netflix as~\cite{Report:ALSH_arXiv14}.

\vspace{-0.05in}
\subsection{Evaluations}\label{sec:hashquality}
 \vspace{-0.05in}

In this section, we show  how the ranking of the two ALSH schemes, L2-ALSH and Sign-ALSH, correlates with the top-$T$ inner products.
Given a user $i$ and its corresponding user vector $u_i$, we compute the top-$T$ gold standard items based on the actual inner products $u_i^Tv_j$, $\forall j$.  We then generate $K$ different hash codes of the vector $u_i$ and all the item vectors $v_j$s and then compute
\begin{equation}
Matches_j = \sum_{t=1}^{K} \mathbbm{1}(h_t(u_i) = h_t(v_j)),
\vspace{-0.05in}
\end{equation}
where $\mathbbm{1}$ is the indicator function and the subscript $t$ is used to distinguish independent draws of $h$. Based on $Matches_j$ we rank all the items. Ideally, for a better hashing scheme, $Matches_j$ should be higher for items having higher inner products with the given user $u_i$.  This procedure generates a sorted list of all the items for a given user vector $u_i$ corresponding to the each hash function under consideration.

For L2-ALSH, we used the same parameters used and recommended in~\cite{Report:ALSH_arXiv14}.  For Sign-ALSH, we used the two recommended choices shown in Section~\ref{sec_parameter}, which are $U=0.75$, $m=2$ and $U=0.85$, $m=3$. It should be noted that Sign-ALSH does not have the parameter $r$.

We compute the precision and  recall of the top-$T$ items for $T\in\{1, 5, 10\}$, obtained from the sorted list based on $Matches$. To compute this precision and recall, we start at the top of the ranked item list and walk down in order. Suppose we are at the $k^{th}$ ranked item, we check if this item belongs to the gold standard top-$T$ list. If it is one of the top-$T$ gold standard item, then we increment the count of \emph{relevant seen} by 1, else we move to $k+1$. By $k^{th}$ step, we have already seen $k$ items, so the \emph{total items seen} is $k$. The precision and recall at that point is then computed as:
\begin{align}
 Precision = \frac{\text{relevant seen}}{k}, \hspace{0.3in}
Recall = \frac{\text{relevant seen}}{T}\notag
\end{align}
We show performance for $K \in \{64,128,256,512\}$. Note that it is important to balance both precision and recall. The method which obtains higher precision at a given recall is superior. Higher precision indicates higher ranking of the relevant items. We  report averaged precisions and recalls over 2000 randomly chosen users.

The plots for MovieLens and Netflix datasets are shown in Figure~\ref{fig_MovielensRanking} and Figure~\ref{fig_NetflixRanking} respectively. We can clearly see, that our proposed Sign-ALSH scheme gives significantly higher precision recall curves than the L2-ALSH scheme,  indicating better correlation of the top neighbors under inner products with Sign-ALSH compared to L2-ALSH.  In addition, there is not much difference in the two different combinations of the parameters $U$ and $m$ in Sign-ALSH. The results are very consistent across both datasets.

\vspace{-0.1in}
\section{LSH Bucketing Experiments}
\vspace{-0.1in}

In this section, we evaluate the actual savings in the number of inner product evaluations for recommending top-$T$ items for the MovieLens dataset. For this, we implemented the standard $(K,L)$ algorithms in~\cite{Proc:Indyk_STOC98}, where $K$ is number of hashes in each hash table and $L$ is the total number of tables. For each query point, the returned results are the union of matches in all $L$ tables.  To find the top-$T$ items, we need to compute the actual inner products only on the candidate items retrieved by the bucketing procedure.

In this experiment, we choose $T\in \{1,5,10\}$ and compute the recall value for each combination of $(T,K,L)$ for every query. For example, given query $q$ and a $(K,L)$-LSH scheme, if $T=10$ and only 5 of the true top-10 data points are retrieved, the recall will be $50\%$ for this $(T,K,L)$. At the same time, we can also compute the \textbf{FIP} (fraction of inner products):
\begin{equation}
\text{FIP}=\frac{(K \times L)  +  TotalRetrieved}{ \text{Total Items}}
\end{equation}
which is basically the total number of inner products evaluation  (where $K\times L$ represents the cost of hashing), normalized by the total number of items in the repository.  Thus, for each $q$ and $(T,K,L)$, we can compute two values: recall and FIP. We also need to figure out a way to aggregate the results for all queries.

Typically the performance of bucketing algorithm is very sensitive to the choice of hashing parameters $K$ and $L$. Ideally, to find best $K$ and $L$, we need to know the operating threshold $S_0$ and the approximation ratio $c$ in advance. Unfortunately, the data and the queries are very diverse and therefore for retrieving top-$T$ near neighbors there is no common fixed threshold $S_0$ and approximation ratio $c$ that works for different queries.

Our goal is to compare the hashing schemes, and minimize the effect of $K$ and $L$ on the evaluation. To get away with the effect of $K$ and $L$, we perform rigorous evaluations of various $K$ and $L$ which includes optimal choices at various thresholds.  For both the hashing schemes, we then select the best performing $K$ and $L$ and report the performance. This involves running the bucketing experiments for thousands of combinations and then choosing the best $K$ and $L$ to marginalize the effect of parameters in the comparisons. This all ensures that our evaluation is fair.

We choose the following scheme. For each $(T,K,L)$, we compute the averaged recall and averaged FIP, over all queries. Then for each ``target'' recall level (and $T$), we can find the $(K,L)$ which produces the best (lowest) averaged FIP. This way, for each $T$, we can compute a ``FIP-recall'' curve, which can be used to compare Sign-ALSH with L2-ALSH. We use $K \in \{4,5,..,20\}$ and $L \in\{1,2,3, ...,200\}$.

The results are summarized in Figure~\ref{fig_MovieLensLSH}.  We can clearly see from the plots that for achieving the same recall for top-$T$, Sign-ALSH scheme needs to do  less computations compared to L2-ALSH.

\vspace{-0.1in}
\section{Conclusion }
\vspace{-0.1in}

The MIPS (maximum inner product search) problem has numerous important applications in machine learning, databases, and information retrieval.  \cite{Report:ALSH_arXiv14}  developed the framework of Asymmetric LSH and provided an explicit scheme (L2-ALSH) for approximate  MIPS in sublinear time. In this study, we present another asymmetric transformation scheme (Sign-ALSH) which converts the problem of maximum inner products into the problem of maximum correlation search, which is subsequently solved by  sign random projections.  Theoretical analysis and experimental study demonstrate that {\em Sign-ALSH} can be noticeably more advantageous than {\em L2-ALSH}.

\vspace{-0.1in}
\section*{Acknowledgement }
\vspace{-0.1in}
The research is supported in part by ONR-N00014-13-1-0764, NSF-III-1360971, AFOSR-FA9550-13-1-0137, and NSF-Bigdata-1419210. The method and  theoretical analysis for {\em Sign-ALSH} were conducted right after the initial submission  of our first work on  ALSH~\cite{Report:ALSH_arXiv14} in February 2014. The intensive experiments (especially the LSH  bucketing experiments), however, were not fully completed until June 2014 due to the demand of computational resources, because we exhaustively experimented a wide range of $K$ (number of hashes) and $L$ (number of tables) for implementing $(K,L)$-LSH schemes. Here, we also would like to thank the computing supporting team (LCSR) at Rutgers CS department as well as the IT support staff at Rutgers Statistics department, for setting up the workstations especially the server with 1.5TB memory.



{

}

\end{document}